\documentclass{ws-ijitdm}
\usepackage[english]{babel}
\usepackage[square,numbers]{natbib}
\usepackage{url}

\usepackage{tabularx} 
\usepackage{hyperref}
\hypersetup{colorlinks=true, linkcolor=blue, filecolor=magenta, urlcolor=red, pdfpagemode=FullScreen,}

\begin{document}
\renewcommand{\footnotesize}{\fontsize{9pt}{11pt}\selectfont}
\title{A New Approach for Multicriteria Assessment in the Ranking of Alternatives Using Cardinal and Ordinal Data}

\author{Fuh-Hwa Franklin Liu \footnotemark{}\footnotetext{Corresponding author; fliu@nycu.edu.tw} }
\address{Professor Emeritus, Department of Industrial Engineering and Management,\\  National Yang Ming Chiao Tung University, Taiwan 300, Republic of China}
\author{Su-Chuan Shih \footnotemark{}\footnotetext{scshih@gm.pu.edu.tw}}
\address{Associate Professor, Department of Business Administration,\\ Providence University, Taiwan 433, Republic of China}

\maketitle
\begin{abstract}
Modern methods for multi-criteria assessment (MCA), such as Data Envelopment Analysis (DEA), Stochastic Frontier Analysis (SFA), and Multiple Criteria Decision-Making (MCDM), are utilized to appraise a collection of Decision-Making Units (DMUs), also known as alternatives, based on several criteria. These methodologies inherently rely on assumptions and can be influenced by subjective judgment to effectively tackle the complex evaluation challenges in various fields. In real-world scenarios, it is essential to incorporate both quantitative and qualitative criteria as they consist of cardinal and ordinal data. Despite the inherent variability in the criterion values of different alternatives, the homogeneity assumption is often employed, significantly affecting evaluations. To tackle these challenges and determine the most appropriate alternative, we propose a novel MCA approach that combines two Virtual Gap Analysis (VGA) models. The VGA framework, rooted in linear programming, is pivotal in the MCA methodology. This approach improves efficiency and fairness, ensuring that evaluations are both comprehensive and dependable, thus offering a strong and adaptive solution. Two comprehensive numerical examples demonstrate the accuracy and transparency of our proposed method. The goal is to encourage continued advancement and stimulate progress in automated decision systems and decision support systems.

\end{abstract}

\section{Multiple Criteria Assessment (MCA) Problems in Practical Applications }\label{sec:1}
In common multiple-criteria assessment (MCA) situations, various criteria are used to evaluate a collection of decision-making units (DMUs). \citet{Sickle} explained in their textbook that both Data Envelopment Analysis (DEA) and Stochastic Frontier Analysis (SFA) serve as the primary methods utilized for tackling MCA issues detailed within decision matrices. \citet{Taherdoost} prescribed the MCA decision matrix analyzed in the multiple-criteria decision-making (MCDM) literature, whereas the columns are identified as \textit{alternatives}, while the rows of performance metrics are classified into two categories: minimization \textit{criteria} and maximization \textit{criteria}. Throughout this article, the phrases "decision-making units" and "alternatives" are used interchangeably.

In evaluations rooted in best (worst) practices, it is typically advantageous for input values to be lower and output values to be higher (or vice versa). The decision matrix features rows titled "input and output metrics," alongside "minimization criteria" and "maximization criteria."
MCA has been extensively utilized across various practical applications, including engineering, social sciences, finance, and general management. While these inputs and outputs might not exactly match the actual production parameters, they are chosen to facilitate a wide-ranging comparative assessment of the DMUs.

To effectively evaluate \textbf{practical applications}, it is important to utilize a blend of both quantitative and qualitative metrics, employing \textbf{cardinal and ordinal data} to generate insights based on data analysis. Utilizing ordinal data is vital as it captures distinct qualitative traits that can be subjectively assessed through techniques like group consensus, surveys, or voting systems, and are typically visualized using Likert scales. Comprehensive discussions on qualitative assessment techniques are provided in the Analytical Hierarchy Process (AHP) \citep{Hansen} and Qualitative Comparative Analysis (QCA) \citep{Scholz}. However, the discrete nature of Likert scale values introduces additional challenges when implementing systematic methods to address related issues.

 Moreover, not all DMUs aim for the same objectives, even if they implement similar procedures or work within an identical environment, resulting in\textbf{ heterogeneous DMUs} within the decision matrix. Typically, the panel of MCA decision-makers relies on implicit knowledge to evaluate these DMUs. These aspects should be considered to ensure accurate evaluations.
 
Existing MCDM methods and models like DEA and SFA fall short when it comes to managing the intricate challenges of MCA. The Virtual Gap Analysis (VGA) model, noted in \citet{Liu2025}, uniquely succeeded in addressing MCA issues without involving ordinal data. 

With an increasing emphasis on automated decision systems (ADS) and decision support systems (DSS), as emphasized by \citet{Heavin} and \citet{Power2019}, we have modified the VGA into the Ord-VGA method to integrate cardinal and ordinal data for MCA. In tackling an MCA issue, a decision matrix consisting of \textit{positive values} requires a comprehensive evaluation via the Ord-VGA method. 

The organization of the paper is subsequently outlined as follows.
\begin{itemize}
\item Section \ref{sec:2} the literature review.
\item Section \ref{sec:3} introduces the two-stage framework of the Ord-VGA to select the best alternative.
    \item Section \ref{sec:4} delves into the Ord-VGA model of Stage I, where we explore the duality features of both primal and dual programs.
    \item Section \ref{sec:5}  focuses on explaining the Ord-VGA model of Stage II.
     \item Section \ref{sec:6}  explorations of the derived optimal solutions.
    \item Section \ref{sec:7}  presents a minimal working example and a larger size real problem to illustrate the benefits of Ord-VGA.
    \item Section \ref{sec:8}  concludes the paper and discusses the key conclusions.
\end{itemize}

\section{Literature Reviews} \label{sec:2}
During the last five decades, DEA models grounded in linear programming, and SFA models based on statistical techniques, methods in MCDM have dominated the field of MCA problem-solving, leading to the production of numerous articles. The deficiencies of the MCDM, DEA, and SFA methods in handling MCA were examined. \citet{Liu2025} recently revealed VGA  models distinctively offered comprehensive evaluations for MCA issues involving cardinal data.

\subsection{Limitations of existing MCDM methods} \label{sec:2.1}
A range of MCDM methods has been developed to address intricate decision-making challenges across various domains. \citet{Sahoo} highlighted the advancements, applications, and future trajectories of MCDM methods and observed that these methods often depend on subjective judgments, rendering them vulnerable to personal biases. 

\citet{Amor} disclosed a map of academic research on multicriteria sorting, classification, and clustering methods and highlighted the key research trends and avenues by conducting a bibliometric analysis. Similarly, \citet{Danielson} investigated the inconsistencies associated with the MCDM ranking techniques. \citet{Danielson2014} suggested integrating methods for improving the expressiveness of user statements.

\citet{Taherdoost} deliver an extensive examination of MCDM, covering its fundamental concepts, uses, main classifications, and approaches, while detailing 18 key MCDM methods. According to \citet{Mardani}, the analytical hierarchy process (AHP) stands as the most prominently used independent MCDM tool, with hybrid MCDM methods following closely. \citet{Keenan} performed a sociometric investigation of MCDM to clarify its development and trends.

\citet{Antunes} developed a hybrid DEA-MCDM technique to assess the effectiveness and synergy of DMUs, enabling their classification according to the performance synergy tiers. \citet{Volaric} combined a Fuzzy Analytic Hierarchy Process (FAHP) with the MCDM method TOPSIS to determine the optimal multimedia application for educational purposes. \citet{Ban} modified an approach in fuzzy multicriteria decision making to identify the most effective DMU.

\citet{Cernev} highlighted the importance of structuring complex evaluation tasks in financial decision-making, utilizing quantitative and qualitative criteria for analysis. This indicates that future research should incorporate machine learning, artificial intelligence, and MCDM models to address complex financial problems effectively.

\subsection{Limitations of Stochastic Frontier Analysis (SFA)} \label{sec:2.2} The work by \citet{Kumbhakar} offers an extensive elucidation of statistic-based SFA methodologies. \citet{Theodoridis} effectively articulated these constraints. Ordinal metrics do not satisfy the statistical conditions requisite for inclusion in SFA models.

\subsection{Limitations of DEA Models Handling Cardinal Data}\label{sec:2.3} 
\citet[in][Chapters 1 and 2]{Sickle} illustrate the assumptions of DEA. In the linear programming DEA models without and with the convexity constraint respectively exhibit constant and variable return to scale, abbreviated as CRS and VRS. Many CRS and VRS \textbf{additive DEA models} have attempted to measure the maximum relative efficiency score of $DMU_o$ via simultaneous adjusting the inputs and outputs values. The modified or extended additive DEA models result in partial estimates. 

\citet{Dyson, Halicka, Mergoni, Liu2023} along with numerous studies have thoroughly investigated the limitations of DEA models. \citet{Liu2023} explored the underlying factors contributing to the inaccuracy of DEA models during their evaluations. 
\subsection{Challenges of Heterogeneity in DEA Models}\label{sec:2.4} All DEA models assume that the DMUs are homogeneous. This assumption is critical for applying an \textit{artificial goal vector of weights}. As noted by \cite{Aleskerov, Zhu, 2016LiW, Zarrin}, which highlights concerns over the composition of DMUs within the decision matrix, in particular, the potential for inputs not to influence all outputs, or for detailed input-output relationships to remain insufficiently examined. The complexity of this issue is underscored by a vast body of research, encompassing tens of thousands of studies conducted over the past five decades. Heterogeneity among DMUs leads to flawed evaluations, and no single method consistently provides reliable results. 

The concerns articulated by \citet{Zarrin} highlight that discrepancies in DEA efficiency scores are attributable to the intrinsic heterogeneity of DMUs and that existing DEA models exhibit a deficiency in predictive capabilities. The proposed framework for evaluating DMU performance employs a synthesis of two complementary modeling techniques: a \textit{self-organizing map artificial neural network} for the purpose of cluster analysis and a multilayer perceptron \textit{artificial neural network} to facilitate heterogeneity analysis and best practice evaluation. 

\subsection{Challenges in DEA Models for Mixed Cardinal and Ordinal Data}\label{2.5} \citet{Ebrahimi:2024} thoroughly analyzed DEA models that included ordinal data. Despite this, many ordinal DEA models cited in \citep{Ebrahimi:2020, Amin:2013, Amin:2022, Chen, Ebrahimi:2024, Toloo} do not ensure accurate determination of efficiency scores. Furthermore, these models lack verification of their duality properties. Without assessing strong complementary slackness conditions (SCSC), solutions from linear programming-based models may be unreliable.

\citet{Chen} employed three DEA models to evaluate the adjustment ratios of three output values, which are presented in three tables \cite[see][Tables 2-4]{Chen}. Each output metric was consistently assigned a weight of one-third. 

\subsection {VGA Models Provides Comprehensive Assessments for Cardinal Data}\label{sec:2.6} \citet{Liu2025} recommended the use of VGA models to identify \textit{virtual gaps} and concurrently adjust the inputs and outputs of $DMU_o$. These models effectively create a \textbf{unified goal price} in virtual currency, setting the baseline for the weighted values of inputs and outputs of $DMU_o$, enabling more comprehensive and accurate analyses. The VGA model's objective value is the virtual gap, defined as the difference between the total weighted inputs and outputs. This virtual gap is always within the range [\$0, \$1), essentially representing a dimensionless inefficiency measure. 

\begin{figure}[ht!] 
\centering
 \includegraphics[width=0.84 \textwidth, height=4 in]{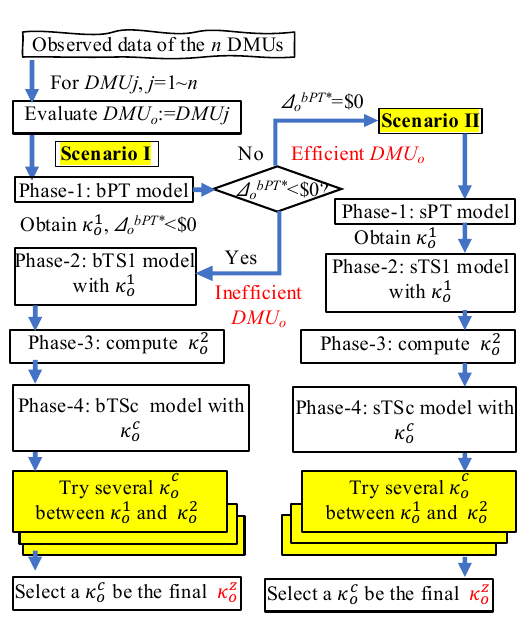} 
 \caption {The framework for the MCA method which utilizes four VGA models. \cite[See] [Figure 1]{Liu2025}} \label{fig1}
  \end{figure}

\citet{Liu2025} explains that Figure \ref{fig1} outlines a framework for the MCA method, which utilizes VGA models to assess virtual gaps of DMUs. The method designed to assess and rank DMUs, incorporating two distinct scenarios. In Scenario I, Phase-1 utilizes the \textbf{bPT} VGA model for \textbf{b}est practices of $DMU_o$, quantifying the \textbf{P}ure virtual gap $\Delta_o^{bPT\star}$ in \textbf{T}echnical processes for input reduction and output expansion. $DMU_o$ is categorized as an efficient unit when $\Delta_o^{bPT\star}=\$0$. 

The \textbf{b}est practice \textbf{T}echnical and \textbf{S}calar \textbf{C}hoice (\textbf{bTSc}) VGA models in Phases-2, 3, and 4 assign Scalars $\kappa_o^1$, $\kappa_o^2$, and $\kappa_o^z$, respectively. Scenario I evaluates the range of $\kappa_o^1$ and $\kappa_o^2$, allowing $DMU_o$ to select the final $\kappa_o^z$ to determine the virtual gap $\Delta_o^{bTSc\star}$, incorporating both technical processes and the scaling of inputs and outputs, as feasible in $DMU_o$'s practice.

In Scenario II, the efficient DMUs identified in Scenario I are evaluated and ranked further. During Phase-1, the Super-Pure Technical (sPT) VGA model calculates the super-virtual gap $\Delta_o^{sPT\star}$ for $DMU_o$, focusing on enlarging inputs and reducing outputs to identify the buffer necessary to sustain efficiency. 

In Phases-2, 3, and 4, the Super Technical and Scalar Choice (sTSc) VGA models assign Scalars $\kappa_o^1$, $\kappa_o^2$, and $\kappa_o^z$ respectively. Scenario II examines the range of $\kappa_o^1$ and $\kappa_o^2$, enabling $DMU_o$ to choose the final $\kappa_o^z$. This choice determines the virtual gap $\Delta_o^{sTSc\star}$, reflecting both technical operations and the scale adjustments of inputs and outputs, in line with $DMU_o$'s practical capabilities.
\vspace{-2 em}  
\subsection{Initial Advances in VGA Models}\label{sec:2.7} \citet{Liu2015} made a substantial impact on the development of Virtual Gap Measurement (VGM) models optimized for use with cardinal data, where the \textit{unified virtual goal price} was determined based on subjective criteria, leading to partial evaluations. Building upon this work, \citet{Liu2017a, Liu2017b} applied the VGM model to two-stage production systems utilizing cardinal datasets.

Professor Shih offered advice on how to incorporate these models into unpublished theses and gave specific assistance for including the bPT model in research projects dealing with cardinal datasets \citep{JCLi2022, LinZY2024}.

\section{Framework of the Novel Two-stage Ord-VGA method}\label{sec:3}
\begin{figure}[ht!] 
\centering
\includegraphics [width=1.04\linewidth, height=1 in]{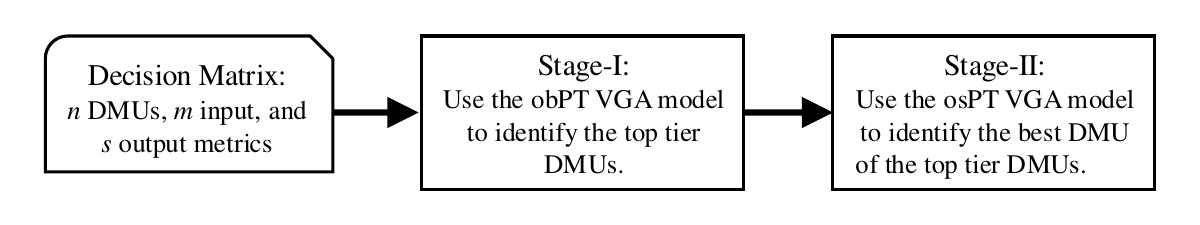}
\caption{The novel MCA method with two stages of Ord-VGA models.} \label{fig2}
 \end{figure}

 The decision matrices for MCA problems comprise both cardinal and ordinal data can be accommodated by the innovative MCA approach that integrates two specific Ord-VGA models, as the framework of Figure \ref{fig2}. The the best-practice Pure Technical (\textbf{obPT}) and the super-virtual gap of Pure Technical (\textbf{osPT}) Ord-VGA models are based on the bPT and sPT VGA models, respectively. The Ord-VGA models are grounded in the central principles of linear programming \citep[see][Sections 1-6]{Bazaraa}, functioning without underlying assumptions. Each Ord-VGA model is comprised of a pair of related programs: primal and dual programs. 
 
As depicted in Figure \ref{fig1}, Phases 2, 3, and 4, the bTSc and sTSc VGA models are currently omitted from the Ord-VGA framework as seen in Figure \ref{fig2}. This is due to the fact that the ordinal data do not represent the technical operations of $DMU_o$; rather, they are assessed by MCA decision-makers. It is essential that $DMU_o$ does not partake in determining the scalar $\kappa_o$, which is used to identify attainable points on the Likert scales. 
 \vspace{-2 em}  
\subsection{Research Principles and Objectives}\label{sec:3.1}
The innovative MCA method, focusing on practical use, is grounded in the following principles.
\begin{enumerate}
\item {Assess DMUs using both quantitative and qualitative criteria, represented by cardinal and ordinal data.}
\item {Translate each qualitative criterion into a Likert Scale through alternative approaches.}
\item {Decision matrices consist of positive values with specific measurement units.}
\item {Acknowledges and accommodates intrinsic \textbf{heterogeneity} among DMUs.}
\item {Develops linear programming-based Ord-VGA models without making extra assumptions.}
\item {Verifies the dual characteristics of every Ord-VGA model.}
\item {Removes subjective biases and personal judgments, without relying on normalization, weighting schemes, or standard criteria across units.}
\item {The computational method is both efficient and effective, quickly identifying the best DMU from decision matrices with numerous DMUs.}
\item {Remove the determined top-performing DMU from the decision matrix and repeat the procedure to find the subsequent best DMU. }
\end{enumerate}

Section \ref{sec:7} provides both a minimal working example and a large-scale real-world example. The thorough solutions demonstrate that this research's objectives have been met.
\begin{enumerate} 
\item This section briefly introduces the fundamental techniques of the new MCA method, where each Ord-VGA model converts the input and output metrics of each DMU into a single \textbf{virtual input} and \textbf{virtual output} pair. 
\item These virtual input and output pairs for the DMUs are illustrated on a two-dimensional intuition graph. 
\item The 2D graph reveals that the virtual gap of $DMU_o$ falls within the interval [\$0, \$1) or the inefficiency score is between [0,1). 
\item Furthermore, the 2D graph demonstrates that with the adjusted input and output values for $DMU_o$, no virtual gap consistently appears. 
\item The 2D graph also allows visualization of the reference peers of $DMU_o$.
\end{enumerate}

\subsection{Notations and Proper Nouns } \label{sec:3.2}
 A collection of $n$ Decision Making Units (DMUs), referred to as $J$, is assessed based on both quantitative and qualitative performance criteria. These criteria are represented as ($I^C, R^C$) for the numerical data, and ($I^O, R^O$) for the data on an ordinal Likert scale. The performance metrics are categorized into \textit{m} inputs, specifically $I^C$ and $I^O$, and \textit{s} outputs, namely $R^C$ and $R^O$. 

For instance, Table \ref{table1} presents the decision matrix of example $(X, Y)$, which consists of row vectors representing inputs and outputs, denoted as $X_i$ where $i=1,...,m$ and $Y_r$ where $r=1,...,s$. For any column related to $DMU_j$, the measured values of $X_i$ and $Y_r$ are articulated as $(x_j, y_j)^t$, where $x_{ij}$ and $y_{rj}$ correspond to particular measurements. The pair of (minimum, maximum) rankings on the Likert scale for $X_i$ and $Y_r$ are indicated by ($M_i^{xL}$ and $M_i^{xU}$) and ($M_r^{yL}$ and $M_r^{yU}$), respectively.

Within the VGA model, every $DMU_o$ was methodically assessed, employing symbols that include the subscript 'o.' 
\begin{itemize}
     \item ($v_{io}$, $u_{ro}$) – \textbf{Virtual unit price} (\$) for ($X_i$, $Y_r$), represented in vector form as $(V_o, U_o)$.
    \item ($d_{io}^x, d_{ro}^y$) – \textbf{Modifications of the virtual unit prices $(\$)$ for Likert scale values} associated with ($X_i$, $Y_r$), with vector expressions ($D_o^x, D_o^y$).
    \item ($q_{io}$, $p_{ro}$) – \textbf{Adjustment ratios} (dimensionless) for ($X_i$, $Y_r$), presented as vectors $(Q_o, P_o)$.
    \item ${\pi_{oj}}$ is the \textbf{intensity variable} (dimensionless) for $DMU_j$ assessed in relation to $DMU_o$, and is shown in vector form as ${\Pi_o}$.
\end{itemize}
  A \textbf{two-step approach} was introduced to address each VGA model. The decision variables are denoted with superscripts '$\#$' and '$\star$', representing the optimal solutions derived in Step I and Step II, respectively. In Step I, the \textbf{unified goal price} $\tau_o^\#$ is set to $\$1$, expressed in virtual currency (\$). The second unified goal price $\tau_o^\star$ for $DMU_o$ is systematically inferred from the results of Step I. 

The primal program aims to measure the virtual gap of $DMU_o$ using variable virtual unit prices $(V_o, U_o)$ modified with ($D_o^x, D_o^y$). Where the product of the value and virtual unit price of each metric is its \textbf{virtual price}. The distinct measurement units of the input and output metrics were converted into a virtual currency $(\$)$. Each \textbf{virtual price} is bounded by the unified goal price. Utilizing a unified goal price is one of the core techniques of the VGA.

Conversely, Step 1 of the dual program optimizes the \textbf{total adjustment price}, the sum of adjustment ratios multiplied $\tau_o^\#$, in terms of the \textit{adjustment ratios} $(Q_o, P_o)$ as well and the \textit{intensities} of the DMUs, ${\Pi_o}$. 

 Furthermore, the aggregation of virtual prices assigned to the inputs and outputs results in the emergence of the \textbf{virtual input} \(\alpha_o^\star\) as well as the \textbf{virtual output} \(\beta_o^\star\). 
 
\subsection{Outlines of Stages I and II}\label{sec:3.3}
In the dual formulation of the \textbf{obPT model}, \textbf{Stage I} systematically identifies each of the inputs and outputs within $DMU_o$, ensuring adherence to the prescribed \textit{reduction and expansion} \textbf{target values}. It is expected that the \textbf{reduced virtual gap} \textbf{$\Delta_o^{obPT\star}$} and the \textbf{maximized overall adjustment cost} \textbf{$\delta_o^{obPT\star}$}, relating to the objective values in both the primal and dual formulations, will coincide. Subsequently, $\Delta_o^{obPT\star}$ is represented as: \( \alpha_o^\star - \beta_o^\star \geq \$0 \). Each of the other DMUs, $DMU_j$, is constrained to possess \( \alpha_{oj}^\star -\beta_{oj}^\star\geq \$0\). The obPT model should be deployed to  comprehensively assess all \textit{n} DMUs. 
  
 Implementation of obPT model to evaluate all \textit{n} DMUs. A $DMU_o$ with $\Delta_o^{obPT\star} = \$0$ is classified as the \textbf{top-tier}. Stage II differentiates the performances of these top-tier DMUs.

In \textbf{Stage II}, the dual program determines the \textit{expansion and reduction} for each input and output of $DMU_o$ in alignment with the target values. The primal program of the \textbf{osPT model} assesses the \textbf{super-virtual gap} $\Delta_o^{osPT\star} $ for each $DMU_o$, expressed as the virtual input deficit relative to the virtual output, \( \alpha_o^\star -\beta_o^\star<\$0\). Whereas each of the remaining top-tier DMUs, $DMU_j$, is subjected to the restriction of \( \alpha_{oj}^\star -\beta_{oj}^\star\geq \$0\). It is crucial that the maximized and minimized objective values of the primal and dual programs, $\Delta_o^{osPT\star}$ and $\delta_o^{osPT\star}$, reach equivalence.
 
 The DMU with the \textbf{minimal super-virtual gap} among the top-tier DMUs demonstrated superior performance compared with the others. In both Stages I and II, the obPT and osPT models delivered precise evaluations, as validated by the subsequent outcomes.

For $DMU_o$, the derived input and output benchmark values achieve a balance where the \textbf{benchmark virtual input} equals the \textbf{benchmark virtual output}, demonstrated by $\widehat{\alpha}_o^\star= \widehat{\beta}_o^\star$. In both Stage I and Stage II, the ratios $(\alpha_o^\star - \beta_o^\star)/\alpha_o^\star$ and $(\beta_o^\star - \alpha_o^\star)/\beta_o^\star$ provide measures for $DMU_o$'s relative inefficiency, with values ranging between 0 and 1. Concurrently, $DMU_j$, identified by $\pi_{oj}^\star>0$ and \( \alpha_{oj}^\star - \beta_{oj}^\star = \$0\), serves as a reference point for $DMU_o$, indicating similar input-output features with $DMU_o$.

\section{Stage I: Identify the Top Tier DMUs}\label{sec:4}
Assess each DMU in the set to identify the highest-performing DMUs that display a virtual gap of zero.

\subsection{The obPT model} \label{sec:4.1} The obPT model represents the abbreviations for \textbf{ordinal} data, \textbf{best} practices, and \textbf{pure} \textbf{technical} measurements. For each $DMU_o$ where $o$ is an element of $J$, primal and dual programs were presented. Every constraint is prefixed by its corresponding dual variable, thus maintaining uniformity in the transformation of the programs. Section \ref{sec:4.6} elaborates on the connections between the primal and dual programs to affirm the strong complementary slackness conditions (SCSC), which verifies the establishment of the models.

Standard form of the \textbf{dual program} pertinent to the obPT model computes the maximum \textit{total adjustment price} (TAP) $\delta_o^{obPT\star}$ for any specified $DMU_o$. The terms within the brackets following the $\Leftrightarrow$ symbol represent the original expressions.

\begin{equation} \begin{aligned} \label{Eq:1}	
    \delta_o^{obPT\star} (\$)= \max_{Q_o , P_o , \Pi_o } \sum_{\forall i\in I^C \cup I^O} q_{io} \tau_o  &+ \sum_{\forall r\in R^C \cup R^O}p_{ro} \tau_o \quad \forall o \in J
    \end{aligned}
 \end{equation}
 \quad \quad subject to:
\begin{equation} \begin{aligned}
 v_{io}: & \sum_{\forall j \in J} x_{ij} \pi_{oj} + q_{ro} x_{io} =  x_{io} \quad\forall i \in I^C \cup I^O	\\
 & \Leftrightarrow [\sum_{\forall j \in J} x_{ij} \pi_{oj}  = (1- q_{ro}) x_{io}]
\label{Eq:2}
\end{aligned}
 \end{equation}
\begin{equation} \begin{aligned}
u_{ro}: &- \sum_{\forall j \in J} y_{rj} \pi_{oj} + p_{ro} y_{ro} = - y_{ro}\quad \forall r \in R^C \cup R^O 	\\
&\Leftrightarrow [\sum_{\forall j \in J} y_{rj} \pi_{oj} =(1+ p_{ro} )y_{ro}]
\label{Eq:3}
\end{aligned}
 \end{equation}
 \begin{equation} 
d_{io}^x:  -(1-q_{io}) x_{io} \leq -M_i^{xL}\Leftrightarrow [(1-q_{io}) x_{io} \geq M_i^{xL}]\quad \forall i \in I^O	
\label{Eq:4}
 \end{equation}
 \begin{equation} 
d_{ro}^y:  (1+p_{ro}) y_{ro}\leq M_r^{yU}\quad  \forall r \in R^O	
\label{Eq:5}
 \end{equation}
\begin{equation} \label{Eq:6}
\Pi_o , Q_o, P_o  \ge 0	
 \end{equation}

In accordance with the principles of linear programming, the dual program was transformed into a \textbf{primal program} that aimed to ascertain the minimum \textit{total virtual gap} (TVG) $\Delta_o^{obPT\star}$ for $DMU_o$. Equation (\ref{Eq:7}) comprises four \textbf{virtual price} categories of $DMU_o$. 

Equation (\ref{Eq:8}) uses virtual prices ($V_o, U_o$) to assess the virtual gap for each $DMU_j$. Certain DMUs, termed \textit{reference peers of $DMU_o$}, have their estimated virtual gaps equate to zero. The values ($V_o^\star, U_o^\star$) represent the estimated unit prices of inputs and outputs used by reference peers in their technical processes to convert minimization inputs into maximization outputs. The \textbf{obPT} model assesses the pure technical virtual gap for $DMU_o$, expressed as ($\Delta_o^{obPT\star}\geq \$0$).

Equations (\ref{Eq:9})–(\ref{Eq:12}) establish a lower boundary for \textit{ virtual prices}, referred to as the \textit{unified goal price} $\tau_o (\$)$. During Step I and Step II, $\tau_o$ in this context is substituted by $\tau_o^\# = \$1$ and $\tau_o^\star = \$ \bar{t}$, respectively.
 \begin{equation}
\begin{aligned}
	&\Delta_o^{obPT\star} (\$)= \min_{V_o, U_o, D_o^x, D_o^y}
    \sum_{\forall i\in I^C} v_{io} x_{io}+ \sum_{\forall i\in I^O} [-M_i^{xL}d_{io}^x +x_{io} (v_{io}+d_{io}^x)] \\ 
    & \qquad \quad\quad  -\{ \sum_{\forall r\in R^C} y_{ro} u_{ro} -\sum_{\forall r\in R^O} [-M_r^{yU}d_{ro}^y +y_{ro}(u_{ro}+d_{ro}^y)]\}\quad \forall o \in J	
 \label{Eq:7}
 \end{aligned}
 \end{equation}
 \quad \quad subject to:
 \begin{equation}	
    \quad \pi_{oj}: \sum_{\forall i\in I^C\cup I^O} v_{io} x_{ij}  - \sum_{\forall r\in R^C \cup R^O}u_{ro}  y_{rj}  \ge 0(\$)\quad \forall j\in J \label{Eq:8}
 \end{equation}
 \begin{equation}
	q_{io}: x_{io} v_{io} \ge\tau_o(\$)\quad \forall i\in I^C	
 \label{Eq:9}
 \end{equation}
 \begin{equation}
	q_{io}: x_{io} (v_{io}+d_{io}^x) \ge\tau_o(\$)\quad \forall i\in I^O	
 \label{Eq:10}
 \end{equation}
\begin{equation}\label{Eq:11}
	p_{ro}: y_{ro} u_{ro}\ge \tau_o(\$)\quad \forall r \in R^C
 \end{equation}
 \begin{equation}\label{Eq:12}
	p_{ro}: y_{ro} (u_{ro}+ d_{ro}^y) \ge \tau_o(\$)\quad \forall r \in R^O
 \end{equation}
\begin{equation}
	V_o, U_o \quad free,\quad  D_o^x, D_o^y \ge 0	
 \label{Eq:13}
 \end{equation}
 \subsection{Obtaining the Virtual Input and Virtual Output of all DMUs}\label{sec:4.2}
Equations (\ref{Eq:30}) and (\ref{Eq:31}) facilitate adherence to the conditions specified in the primal program. The solutions associated with Equation (\ref{Eq:1}) for Steps I and II are presented in Equation (\ref{Eq:14}), and Equation (\ref{Eq:15}) delineates the solutions pertaining to Equation (\ref{Eq:7}), which is relevant to the aforementioned steps. Within Equation (\ref{Eq:15}), the four virtual gap prices are illustrated as pairs, comprising virtual input (\textbf{vInput}) and virtual output (\textbf{vOutput}), represented by $(\alpha_o^\#, \beta_o^\#)$ and $(\alpha_o^\star, \beta_o^\star)$. During Step I, the comprehensive virtual gap, $\Delta_o^{obPT\#}$ can exceed \$1. The cumulative virtual gap consists of four distinct components.
 \begin{equation}
 \begin{aligned}
\$0 \leq &\delta_o^{obPT\#} (\$)= \delta_{xo}^\# +\delta_{yo}^\#
= \sum_{\forall i\in I^C \cup I^O} q_{io} ^\# \times \$1  + \sum_{\forall r\in R^C \cup R^O}p_{ro} ^\#  \times \$1\\ 
\$0 \leq & \delta_o^{obPT\star } (\$)= \delta_{xo}^\star +\delta_{yo}^\star
=\sum_{\forall i\in I^C \cup I^O} q_{io}^\star  \tau_o ^\star  + \sum_{\forall r\in R^C \cup R^O} p_{ro}^\star  \tau_o ^\star < \$1
  \end{aligned}  \label{Eq:14} 
 \end{equation}  
 \begin{equation} \label{Eq:15}
\begin{aligned}
\$0 \leq & \Delta_o^{obPT\#}(\$) =\{vInput^\#\}-\{vOutput^\#\} = \alpha_o^\# - \beta_o^\#\\
& \qquad \quad =\{\sum_{\forall i\in I^C} x_{io} v_{io}^\# 
    +\sum_{\forall i\in I^O} [-M_i^{xL}d_{io}^{x\#} +x_{io} (v_{io}^\#+d_{io}^{x\#})]\} \\ 
    & \qquad \quad\quad  -\{ \sum_{\forall r\in R^C} y_{ro} u_{ro}^\# -\sum_{\forall r\in R^O} [-M_r^{yU}d_{ro}^{y\#} +y_{ro}(u_{ro}^\#+d_{ro}^{y\#})]\}\\
\$0 \leq &\Delta_o^{obPT\star }(\$)=\{vInput^\star\}-\{vOutput^\star\}=\alpha_o^\star -\beta_o^\star\\
& \qquad \quad =\{\sum_{\forall i\in I^C} x_{io} v_{io}^\star 
    +\sum_{\forall i\in I^O} [-M_i^{xL}d_{io}^{x\star} +x_{io} (v_{io}^\star+d_{io}^{x\star})]\} \\ 
    & \qquad \quad\quad  -\{ \sum_{\forall r\in R^C} y_{ro} u_{ro}^\star -\sum_{\forall r\in R^O} [-M_r^{yU}d_{ro}^{y\star} +y_{ro}(u_{ro}^\star+d_{ro}^{y\star})]\} < \$1
\end{aligned}
 \end{equation}

Equation (\ref{Eq:16}) illustrates the solutions to Equation (\ref{Eq:8}) of each $DMU_j$ during Steps I and II are combined into the pairs (\textit{vInput, vOutput}), specifically $(\alpha_{oj}^\#, \beta_{oj}^\#)$ and $(\alpha_{oj}^\star, \beta_{oj}^\star)$. Notably, in Step I, $\Delta_{oj}^{obPT\#}$ exceeds \$1. 

\begin{equation} \label{Eq:16}
\begin{aligned}
\$0 \leq \quad & \Delta_{oj}^{obPT\#} (\$)= V_o^\# x_j -U_o^\# y_j = \alpha_{oj}^\# - \beta_{oj}^\# \quad \forall j \in J\neq o\\
\$0 \leq &\quad \Delta_{oj}^{obPT\star} (\$)= V_o^\star x_j -U_o^\star y_j = \alpha_{oj}^\star -\beta_{oj}^\star < \$1 \quad \forall j \in J\neq o 
\end{aligned}
 \end{equation}

\subsection{Establish a Unified Goal Price for \texorpdfstring{$DMU_o$}.}  \label{sec:4.3}
In Step II, it is essential for $DMU_o$ to confirm that $\delta_o^{obPT\star}$ and $\Delta_o^{obPT\star}$ fall within the ($\$0, \$1$) interval. The connections between the obPT models in Steps I and II are outlined below.
\begin{equation}\begin{aligned} \label{Eq:17}
\tau_o^\# :\tau_o^\star   &=
 1 :  \bar{t} =  \delta_o^{obPT\#}:\delta_o^{obPT\star} =\Delta_o^{obPT\#}:\Delta_o^{obPT\star}\\
 &= (\alpha_o^\#- \beta_o^\# ) : (\alpha_o^\star-\beta_o^\star)
 \end{aligned} \end{equation}
Thus, the \textit{relation equation} below holds.
\begin{equation} \label{Eq:18}
\bar{t} \$ (   \alpha_o^\#- \beta_o^\# ) = 1\$(\alpha_o^\star-\beta_o^\star) 
\end{equation}
Equation (\ref{Eq:19}) was utilized to derive the dimensionless quantity of $\bar{t}$.
\begin{equation} \label{Eq:19}
	\$\bar{t} = \$1/\alpha_o^\# \textrm {  and  } \tau_o^\star   = \$ \bar{t} 	
  \end{equation}
Equation (\ref{Eq:18}) can be rewritten as Equation (\ref{Eq:20}).
\begin{equation} \begin{aligned} \label{Eq:20} 
 (\$1/\alpha_o^\#) \$(   \alpha_o^\#- \beta_o^\# ) = 1\$(\alpha_o^\star-\beta_o^\star) \Leftrightarrow 
  \$(   1- \beta_o^\#/\alpha_o^\# ) = \$(\alpha_o^\star-\beta_o^\star)
 \end{aligned} \end{equation}
  Dividing Equation (\ref{Eq:20}) by $\alpha_o^\star$, it becomes:
 \begin{equation}\label{Eq:21}
  (\$1/\alpha_o^\star)( 1- \beta_o^\#/\alpha_o^\# ) =(1-\beta_o^\star/\alpha_o^\star )
  \end{equation}
  Equation (\ref{Eq:15}) showed $ \$0 \leq( \alpha_o^\#- \beta_o^\# )$ and $ \$0\leq (\alpha_o^\star-\beta_o^\star) $, which are equivalent to the following equations:
  \begin{equation} \label{Eq:22}
  0\leq( 1- \beta_o^\#/\alpha_o^\# )\leq 1 \text{ and } 0\leq (1-\beta_o^\star/\alpha_o^\star )\leq 1
  \end{equation}
The value $\alpha_o^\star = \$1$ found in Equation (\ref{Eq:21}) verifies the expressions contained in Equation (\ref{Eq:22}), with $\Delta_o^{obPT\star}$ expected to lie between ($\$0, \$1$). In most scholarly discussions, efficiency analysis hinges on directly computing the \textit{efficiency score}. Instead, we propose the computation of the \textit{inefficiency score}, which, when summed with the \textit{efficiency score}, should yield a total of 1. 

Equation (\ref{Eq:23}) illustrates that in Steps I and II, the \textit{virtual inefficiency scores}, represented as  $F_o^{obPT\star}$, are calculated independently of the unified virtual price are restricted within the interval (0, 1). In Step I, each $DMU_o$ utilizes $\tau_o^\#=\$1$ in an unbiased fashion. The objective evaluation process in these two steps identifies $\tau_o^\star$. To clarify, the obPT model evaluates the performance of $DMU_o$ through a two-step process, as described in Equation (\ref{Eq:23}), guaranteeing that $\$0 \leq \Delta_o^{obPT\star} < \$1.$
  \begin{equation}
\begin{aligned}\label{Eq:23} 
	0\leq F_o^{obPT\star}=\Delta_o^{obPT\star}/\alpha_o^\star
    =  \beta_o^\#/\alpha_o^\#= \beta_o^\star/\alpha_o^\star <1  
\end{aligned}
\end{equation}
In mathematical optimization, a common method involves normalizing the Step I solutions to obtain the Step II solutions. In mathematical optimization, a common method involves normalizing the Step I solutions to obtain the Step II solutions. 
\begin{equation} \begin{aligned}
 (  Q_o^\star, P_o^\star,  \Pi_o^\star,)&= ( Q_o^\#, P_o^\#,  \Pi_o^\#)\\
  (\tau_o^\star, \delta_o^{obPT\star },  \Delta_o^{obPT\star }, & V_o^\star, U_o^\star, D_o^{x\star}, D_o^{y\star})\\
  =\bar{t} (\tau_o^\#,\delta_o^{obPT\#}, \Delta_o^{obPT\#}, & V_o^\#, U_o^\#,  D_o^{x\#}, D_o^{y\#})
\label{Eq:24} \end{aligned}
\end{equation}
  In Equation \ref{Eq:15}, the calculated virtual gap for $DMU_o$, specifically $\Delta_o^{obPT\star}$ in Step II, is standardized and confined to the interval [\$0, \$1). This facilitates comparisons between the \textit{n} assessments of different $DMU_o$ by employing the normalized results.  
\subsection{The Top-tier DMUs}\label{4.4}
The symbol $\mathcal{E}_o^{obPT}$ represents the group of \textit{reference peers} for $DMU_o$ in obPT evaluations. A $DMU_j$ is a part of $\mathcal{E}_o^{obPT}$ if it is efficient and satisfies $\pi_{oj}^\star>0$. Conversely, for other inefficient $DMU_j$, it holds that $\pi_{oj}^\star =0$.

When evaluating $DMU_o$, we determined the set of reference peers $\mathcal{E}_o^{obPT}$. The collective group of top-tier DMUs comprises the union of the \textit{n} reference peer sets, represented by the following equation:
\begin{equation} \label{Eq:25}
    \mathcal{E}^{obPT}= \cup_{\forall o \in J} \mathcal{E}_o^{obPT}
\end{equation}

\subsection{Duality Properties}\label{sec:4.5}
We formulated Ord-VGA models based on linear programming theory \citep{Bazaraa}, which requires verification of the duality between each pair of primal and dual programs. Constraints and decision variables must be consistently defined with appropriate measurement units in both programs.

When the optimal values for \textit{total virtual gap} (TVG) and \textit{total adjustment price} (TAP) in the obPT model match, as described in Equation (\ref{Eq:26}) signifies a balance between the specified conditions and goals of the model. This equivalence shows synchronization between the TVG and TAP factors, illustrating a cohesive link between the gap and price measures.
\begin{equation}\begin{aligned} \label{Eq:26}
\delta_o^{obPT\star}(\$)=\Delta_o^{obPT\star} (\$)
  \end{aligned} \end{equation}

The Ord-VGA model, which relies on linear programming, repeatedly goes through multiple steps to identify the goals of both primal and dual programs. It evaluates the optimized results such as ($V_o^\star, U_o^\star$), ($Q_o^\star, P_o^\star$), ($D_o^{x\star}, D_o^{y\star}$) and $\Pi_o^\star$. In addition, it is necessary to confirm the strong complementary slackness conditions of these programs.
 
\subsection{Strong Complementary Slackness Conditions (SCSC) of the obPT model}\label{sec:4.6}
The optimal solutions of Equations (\ref{Eq:1})-(\ref{Eq:4}) are expressed as Equations (\ref{Eq:27}), (\ref{Eq:28}), and (\ref{Eq:29}), respectively. 
\begin{equation} \begin{aligned} \label{Eq:27}
\relax [\sum_{\forall j\in \mathcal{E}_o^{obPT}} x_{ij} \pi_{oj}^\star - x_{io} (1 - q_{io}^\star)] \times v_{io}^\star =\$0 \quad \forall i\in I^C \cup I^O  
\end{aligned}  \end{equation}
\begin{equation} \begin{aligned}
\relax [\sum_{\forall j\in \mathcal{E}_o^{obPT}} y_{rj}\pi_{oj}^\star - y_{ro}(1+p_{ro}^\star )] \times u_{ro}^\star =\$0 \quad 
 \forall r\in R^C \cup R^O
 \label{Eq:28} \end{aligned} \end{equation}
 \begin{equation} \begin{aligned}\label{Eq:29}
    [(1-q_{io}^\star) x_{io}-M_i^{xL}]  \times d_{io}^{x\star}&=\$0\quad \forall i \in I^O\\ [M_r^{yU}-(1+p_{ro}^\star) y_{ro}] \times d_{ro}^{y\star}&=\$0\quad \forall r \in R^O  
\end{aligned}\end{equation} 
\indent The optimal solutions of Equations (\ref{Eq:8})-(\ref{Eq:12}) satisfy the criteria indicated in Equation (\ref{Eq:30}) and Equation (\ref{Eq:31}), respectively.
\begin{equation}
	(V_o^\star x_j -U_o^\star y_j ) \times \pi_{oj}^\star = \$0\quad \forall j\in J 
	\label{Eq:30}
\end{equation} 
\begin{equation}\begin{aligned}\label{Eq:31} 
	(v_{io}^\star x_{io} - \tau_o^\star)  \times q_{io}^\star &= \$0\quad \forall i \in I^C\\
    \quad [(v_{io}^\star +d_{io}^{x\star}) x_{io} - \tau_o^\star ]  \times q_{io}^\star &= \$0\quad \forall i \in I^O \\
    (u_{ro}^\star y_{ro} - \tau_o^\star) \times  p_{ro}^\star&= \$0\quad \forall r\in R^C\\\quad [(u_{ro}^\star+d_{ro}^{y\star}) y_{ro} - \tau_o^\star ] \times p_{ro}^\star&= \$0\quad \forall r\in R^O 
   \end{aligned} \end{equation}
 In Equations (\ref{Eq:27})-(\ref{Eq:31}), one of the two terms on the left-hand side must be equal to zero to render their product zero. The SCSC set returned the exact solutions of the Ord-VGA model. Take Equation (\ref{Eq:27}) as an example; the linear combinations of its reference peers match the adjusted input-i value as expressed by $\relax [\sum_{\forall j\in \mathcal{E}_o^{obPT}} x_{ij} \pi_{oj}^\star = x_{io} (1 - q_{io}^\star)]$, while ensuring $v_{io}^\star\geq 0$. 
 
 In Equation (\ref{Eq:29}), the expression $[M_r^{yU}-(1+p_{ro}^\star) y_{ro}]$ being greater than or equal to zero corresponds to $d_{ro}^{y\star}=0$ and $d_{ro}^{y\star}>0$, respectively. Similarly, for Equation (\ref{Eq:31}), the situation described by [$(u_{ro}^\star+d_{ro}^{y\star}) y_{ro} - \tau_o^\star$] being more than or equal to zero relates to $p_{ro}^\star=0$ and $p_{ro}^\star>0$.

If $(M_r^{yU} > y_{ro})$ and the condition $[M_r^{yU} - (1 + p_{ro}^\star) y_{ro} > 0]$ is met, then it is essential that $d_{ro}^{y\star}=0$. In Equation (\ref{Eq:15}), the expression $[(-M_r^{yU} d_{ro}^{y\star} + y_{ro}(u_{ro}^\star+d_{ro}^{y\star})]$ simplifies to $u_{ro}^\star y_{ro}$. The modified form $[(1 + p_{ro}^\star) y_{ro} > 0]$ may not be exactly equal to an integer, but should fall between two adjacent Likert points. This approximation on the subjective Likert scale has no impact on calculating the virtual gap $\Delta_o^{obPT\star}$. 

A special scenario occurs when the Likert scale of output-r matches or adjusts to the maximum, $M_r^{yU}=y_{ro}$ or $[M_r^{yU} = (1 + p_{ro}^\star) y_{ro}]$. In Equation (\ref{Eq:15}), the virtual price of output-r is tweaked by the penalty virtual price $-M_r^{yU}d_{ro}^{y\star}$. A similar analysis holds for the situation in which input-i has $M_i^{xL} =x_{io}$, or is adjusted as $M_i^{xL} = (1 - q_{io}^\star) x_{io}$.

\subsection{Sensitivity Analysis} \label{sec:4.7}

Within the TVG framework, $V_o$ and $U_o$ represent variables related to the free indices. In both Equations (\ref{Eq:27}) and (\ref{Eq:28}), the left-hand expressions equate to zero, which implies \(V_o^\star >0, U_o^\star >0\). Given that \( (\textit{X, Y}), (D_o^x, D_o^y)>0\) and \(\tau_o^\star >\$0\), Equation (\ref{Eq:31}) verifies that the calculated virtual unit prices confirm \(V_o^\star >0,U_o^\star >0\). The obPT model provides reliable approximations of $p_{io}^\star$ and $q_{ro}^\star$, ensuring that $DMU_o$ has $\alpha_o^\star$ equals $\$1$.

All $DMU_j$ values adhere to the conditions outlined in Equation (\ref{Eq:30}). If $(V_o^\star x_j - U_o^\star y_j) = 0$ and \(\pi_{oj}^\star > 0\), then $DMU_j$ is considered part of \(\mathcal{E}_o^{obPT}\). Section \ref{sec:7.2} illustrates that the \textit{best-practice equator} is the diagonal line at the origin, characterized by \((V_o^\star x_j - U_o^\star y_j) = \$0\) for every $j$ in \(\mathcal{E}_o^{obPT}\). When $(V_o^\star x_o - U_o^\star y_o) = 0$, $DMU_o$ achieves a zero total virtual gap, denoted as $\Delta_o^\star = \$0$, with $\pi_o^{\star} = 1$. 

If $DMU_j$ satisfies the conditions $(V_o^\star x_j - U_o^\star y_j) > 0$ and \(\pi_{oj}^\star =0\), it is excluded from \(\mathcal{E}_o^{obPT}\). Section \ref{sec:7.2} demonstrates that $DMU_j$ has an overall positive virtual gap, represented by $\Delta_{oj}^\star > \$0$, while $\pi_o^{\star} = 0$. Consequently, $DMU_j$ is positioned below the equator. Excluding $DMU_j$ from the decision matrix did not affect the assessment of $DMU_o$.

\section{Stage II: Choose the Best Alternative (DMU)} \label{sec:5} The osPT model is applied to assess the super-virtual gap for each DMU located in the uppermost tier. Then 

 \subsection{The osPT Model}\label{sec:5.1} 
The osPT Ord-VGA model was applied to assess the \textbf{super-virtual gap} for each $DMU_o, \forall o\in \mathcal{E}^{obPT}$, as described in Equation (\ref{Eq:25}).  \footnote{ \citet[][Section 3.5]{Emrouznejad} provide a review of the literature on DEA SBM models that incorporate super-efficiency, as it pertains to the efficiency frontier of set $J$ excluding $DMU_o$.  }

Each constraint starts with its associated dual variable, ensuring consistency across program transformations. The relationships between the prime and dual programs are detailed in Section \ref{sec:5.4} for strong complementary slackness conditions (SCSC). 
 \textbf{The dual program aims to obtain the total adjustment price (TAP).}
\begin{equation} \begin{aligned} \label{Eq:32}	
    \delta_o^{osPT\star} (\$)&= \min_{Q_o, P_o, \Pi_o} \sum_{\forall i\in I^C \cup I^O} q_{io} \tau_o  + \sum_{\forall r\in R^C \cup R^O }p_{ro} \tau_o \quad \forall o \in \mathcal{E}^{obPT}
    \end{aligned}
 \end{equation}
\quad \quad subject to:
 \begin{equation} \begin{aligned} 
 \quad v_{io}: &-\sum_{\forall j \in \mathcal{E}^{obPT} \neq o} x_{ij} \pi_{oj} + q_{ro} x_{io} \geq  -x_{io} \quad  \forall i \in I^C \cup I^O	\\
  &\Leftrightarrow  [\sum_{\forall j \in \mathcal{E}^{obPT} \neq o} x_{ij} \pi_{oj}  \leq (1+ q_{ro}) x_{io}]
\label{Eq:33}
\end{aligned}
 \end{equation}
\begin{equation} \begin{aligned}
u_{ro}: & \sum_{\forall j \in \mathcal{E}^{obPT} \neq o} y_{rj} \pi_{oj} + p_{ro} y_{ro} \geq  y_{ro} \quad \forall r \in R^C \cup R^O\\
& \Leftrightarrow  [\sum_{\forall j \in \mathcal{E}^{obPT} \neq o} y_{rj} \pi_{oj} \geq (1-p_{ro}) y_{ro} ] 	
\label{Eq:34}
\end{aligned}
 \end{equation}
 \begin{equation} 
d_{io}^x:  -(1+q_{io}) x_{io} \geq -M_i^{xU}\Leftrightarrow [(1+q_{io}) x_{io} \leq M_i^{xU}]\quad \forall i \in I^O	
\label{Eq:35}
 \end{equation}
 \begin{equation} 
d_{ro}^y:  (1-p_{ro}) y_{ro}\geq M_r^{yL}\quad \forall r \in R^O	
\label{Eq:36}
 \end{equation}  
\begin{equation} \label{Eq:37}
\Pi_o, Q_o  , P_o  \ge 0	
 \end{equation}
\indent \textbf{The primal program aims to have the total virtual gap (TVG):}  
 \begin{equation}
\begin{aligned}
	\Delta_o^{osPT\star} (\$) & = \max_{ V_o, U_o,  D_o^x,  D_o^y} -\lbrace\sum_{\forall i\in I^C} x_{io} v_{io} 
    +[\sum_{\forall i\in I^O} [M_i^{xU}d^x_{io} + x_{io}(v_{io}-d^x_{io})]\rbrace\\
    &+ \sum_{\forall r\in R^C} y_{ro}u_{ro}  +[\sum_{\forall r\in R^O} [M_r^{yL} d^y_{ro}+ y_{ro}(u_{ro}-d^y_{ro})] \quad \forall o \in \mathcal{E}^{obPT}	
 \label{Eq:38}
 \end{aligned}
 \end{equation}
 \quad \quad subject to:
 \begin{equation}\begin{aligned}
     \quad \pi_{oj}: - &\sum_{\forall i\in I^C \cup I^O} v_{io} x_{ij}  + \sum_{\forall r\in R^C \cup R^O}u_{ro}  y_{rj}  \leq 0(\$)\quad \forall j\in  \mathcal{E}^{obPT} \neq o\\
     &(=\quad\Delta_{oj}^{osPT\star})\label{Eq:39}
 \end{aligned}\end{equation}
 \begin{equation}
	q_{io}: x_{io} v_{io} \leq\tau_o(\$)\quad \forall i\in I^C	
 \label{Eq:40}
 \end{equation} 
 \begin{equation}
	q_{io}: x_{io} (v_{io}-d_{io}^x) \leq\tau_o(\$)\quad \forall i\in I^O	
 \label{Eq:41}
 \end{equation}
\begin{equation}\label{Eq:42}
	p_{ro}: y_{ro} u_{ro}\leq \tau_o(\$)\quad \forall r \in R^C \end{equation}\begin{equation}\label{Eq:43}
	p_{ro}: y_{ro} (u_{ro}- d_{ro}^y) \leq \tau_o(\$)\quad \forall r \in R^O
 \end{equation}
\begin{equation}
	 V_o, U_o, D_o^x, D_o^y \ge 0	
 \label{Eq:44}
 \end{equation}

\subsection {Obtaining the Virtual Input and Virtual Output of all top-tier DMUs} \label{sec:5.2} Similar to Equations (\ref{Eq:14}) and (\ref{Eq:15}), the signs of $\alpha_o^\star$ and $\beta_o^\star$ are reversed. As illustrated in Equation (\ref{Eq:45}), $DMU_o$ represents a negative super-virtual gap. For example, Figure \ref{fig4} illustrates that $DMU_o$ is located above the prime meridian, characterized by a super-virtual gap that serves as the buffer when point H declines to point T.
\begin{equation} \label{Eq:45}
\begin{aligned}
&\$0 \geq  \Delta_{o}^{osPT\#}(\$) =-\{vInput^\#\}+\{vOutput^\#\} = -\alpha_o^\# + \beta_o^\#\\
&\$0 \geq \Delta_{o}^{osPT\star }(\$)=-\{vInput^\star\}+\{vOutput^\star\}=-\alpha_o^\star +\beta_o^\star\\
&\qquad =-\{\sum_{\forall i\in I^C} x_{io} v_{io}^\star 
    +\sum_{\forall i\in I^O} [M_i^{xU}d_{io}^{x\star}+ x_{io}(v_{io}^\star-d_{io}^{x\star})]\}\\
    &\qquad \quad +\{ \sum_{\forall r\in R^C} y_{ro}u_{ro}  + \sum_{\forall r\in R^O} [M_r^{yL} d_{ro}^{y\star}+ y_{ro}(u_{ro}^\star- d_{ro}^{y\star})]\} > -\$1
\end{aligned}
 \end{equation}
Equation (\ref{Eq:46}) consolidates the solutions from Equation (\ref{Eq:39}) for each $DMU_j$ during Step I and Step II into pairs virtual input and virtual output, labeled as (\textit{vInput, vOutput}), specifically $(\alpha_{oj}^\#, \beta_{oj}^\#)$ and $(\alpha_{oj}^\star, \beta_{oj}^\star)$. During Step I, $\Delta_{oj}^{osPT\#}$ can potentially surpass $\$1$. In Figure \ref{fig4}, point B is depicted below the prime meridian, indicating a virtual deficiency that must be enhanced toward the prime meridian. 
\begin{equation} \label{Eq:46}
\begin{aligned}
\$0 \leq &= \Delta_{oj}^{osPT\#} (\$)= V_o^\# x_j -U_o^\# y_j = \alpha_{oj}^\# - \beta_{oj}^\# \quad \forall j \in \mathcal{E}^{obPT} \neq o\\
\$0 \leq &= \Delta_{oj}^{osPT\star} (\$)= V_o^\star x_j - U_o^\star y_j = \alpha_{oj}^\star -\beta_{oj}^\star < \$1 \quad \forall j \in \mathcal{E}^{obPT} \neq o
\end{aligned}
 \end{equation}
 
\subsection {Determine a Unified Goal Price of \texorpdfstring{$DMU_o$}.} \label{sec:5.3}
$DMU_o$ must ensure that $\delta_o^{osPT\star}$ and $\Delta_o^{osPT\star}$ in Step II fall within the range of ($\$0, \$1$). Normalizing these optimal decision variables is crucial for their comparison with the evaluation results of other $DMU_o$.

The solutions of Steps I and II have the following relationships.
\begin{equation} \begin{aligned}\label{Eq:47}
\tau_o^\# :\tau_o^\star   &=
 1 :  \bar{t} =  \delta_o^{osPT\#}:\delta_o^{osPT\star}  =\Delta_o^{osPT\#}:\Delta_o^{osPT\star}\\
 &= (   -\alpha_o^\#+ \beta_o^\# ) : (-\alpha_o^\star+\beta_o^\star)
 \end{aligned}\end{equation}
Therefore, the next \textit{relation equation} is true.
\begin{equation} \label{Eq:48}
\bar{t} \$ ( - \alpha_o^\#+ \beta_o^\# ) = 1\$(-\alpha_o^\star+\beta_o^\star) 
\end{equation}
We used Equation (\ref{Eq:39}) to determine the dimensionless value of $\bar{t}$. 
\begin{equation} \label{Eq:49}
	\$\bar{t} = \$1/\beta_o^\# \textrm {  and  } \tau_o^\star   = \$ \bar{t} 	
  \end{equation}
Equation (\ref{Eq:48}) can be rewritten as Equation (\ref{Eq:50}).
\begin{equation}\begin{aligned} \label{Eq:50} 
 (\$1/\beta_o^\#) \$( -  \alpha_o^\#+ \beta_o^\# ) &= 1\$(-\alpha_o^\star+\beta_o^\star)\quad\\
 or \quad
 \$( - \alpha_o^\#/\beta_o^\#+  1 ) &= \$(-\alpha_o^\star+\beta_o^\star)
 \end{aligned}\end{equation}
  Dividing Equation (\ref{Eq:50}) by $\beta_o^\star$, it becomes:
 \begin{equation}\label{Eq:51}
  (\$1/\beta_o^\star)( - \alpha_o^\#/\beta_o^\#+  1 ) = (-\alpha_o^\star/\beta_o^\star+1)
  \end{equation}
  Equation (\ref{Eq:45}) shows $ \$0 \leq( -\alpha_o^\#+ \beta_o^\# )$ and $ \$0\leq (-\alpha_o^\star+\beta_o^\star) \leq \$1$, we have the relation equations:
  \begin{equation} \label{Eq:52}
  0\leq( - \alpha_o^\#/\beta_o^\#+  1 )\leq 1 \text{ and } 0\leq (-\alpha_o^\star/\beta_o^\star+1)\leq 1
  \end{equation}
  Consequently, by setting $\beta_o^\star = \$1$ in Equation (\ref{Eq:51}) implies that $\alpha_o^\star \leq \$1$, confirming that Equation (\ref{Eq:52}) falls within the interval (0, 1). It is expected that $\Delta_o^{osPT\star}$ will remain within the range of ($\$0, \$1$). According to Equation (\ref{Eq:53}), the \textit{super-inefficiency score} $F_o^{osPT\star}$ does not depend on the unified goal price. 
\begin{equation}
\begin{aligned}\label{Eq:53} 
	0 \leq F_o^{osPT\#}&=F_o^{osPT\star}  =\alpha_o^\#/\beta_o^\#= \alpha_o^\star/\beta_o^\star < 1
\end{aligned}
\end{equation}

In this analysis, the emphasis was on assessing the super-virtual gap instead of the super-inefficiency score. The super-virtual gap for $DMU_H$, given by \(\Delta_{HH}^{osPT\star} = -\alpha_H^\star + \beta_H^\star\), falls within ($\$0, -\$1$), depicting the direct distance for $DMU_H$ to retreat to the prime meridian at point T.

When assessing $DMU_o$, $DMU_j$ is included in the set of reference peers $\mathcal{E}_o^{osPT}$ if $\Delta_{oj}^{obPT\star}=\$0$ and $\pi_{oj}^{osPT\star}>0$. In contrast, if $\Delta_{oj}^{osPT\star}>\$0$ and $\pi_{oj}^{osPT\star}=0$, then $DMU_j$ is not included in $\mathcal{E}_o^{osPT}$. Similar to Equation (\ref{Eq:24}), Equation (\ref{Eq:54}) illustrates the normalization process.
\begin{equation} \begin{aligned}
 (  Q_o^\star, P_o^\star,  \Pi_o^\star)&= ( Q_o^\#, P_o^\#,  \Pi_o^\#)\\
  (\tau_o^\star, \delta_o^{osPT\star },  \Delta_o^{osPT\star }, & V_o^\star, U_o^\star, D_o^{x\star}, D_o^{y\star})\\
  =\bar{t} (\tau_o^\#,\delta_o^{osPT\#}, \Delta_o^{osPT\#},  & V_o^\#, U_o^\#,  D_o^{x\#}, D_o^{y\#})
\label{Eq:54} \end{aligned}
\end{equation}
 In Equation (\ref{Eq:45}), the calculated super-virtual gap for $DMU_o$, specifically $\Delta_o^{osPT\star}$ in Step II, is standardized and confined to the interval [\$0, -\$1). This aids in comparing the evaluations of various $DMU_o$ within the top-tier set $\mathcal{E}^{osPT}$ through the use of standardized outcomes.
\subsection{Strong Complementary Slackness Conditions (SCSC) of the osPT model}\label{sec:5.4}
The osPT framework exhibits dual characteristics similar to those of the obPT framework; however, it differs in that the input and output metrics are interchanged. The optimal solutions of the constraints in the TAP program are outlined in Equations (\ref{Eq:55})–(\ref{Eq:57}).
\begin{equation} \begin{aligned} \label{Eq:55}
\relax [\sum_{\forall j \in \mathcal{E}^{osPT}} x_{ij} \pi_{oj}^\star - x_{io} (1 + q_{io}^\star)] \times  v_{io}^\star =\$0\quad
\forall i\in I^C \cup I^O  
\end{aligned}  \end{equation}
\begin{equation} \begin{aligned}
\relax [\sum_{\forall j\in \mathcal{E}^{osPT}} y_{rj}\pi_{oj}^\star - y_{ro}(1-p_{ro}^\star )] \times u_{ro}^\star =\$0 \quad \forall r\in R^C \cup R^O
 \label{Eq:56} \end{aligned} \end{equation}
 \begin{equation} \begin{aligned}\label{Eq:57}
    [M_i^{xU}-(1+q_{io}^\star) x_{io}]  \times d_{io}^{x\star}&=\$0\quad \forall i \in I^O\\ [(1-p_{ro}^\star)y_{ro}-M_r^{yL}]  \times d_{ro}^{y\star}&=\$0\quad \forall r \in R^O  
\end{aligned}\end{equation} 
Equations (\ref{Eq:58}) and (\ref{Eq:59}) are optimal solutions for the conditions in the TVG program.
\begin{equation}
	(V_o^\star x_j -U_o^\star y_j ) \times \pi_{oj}^\star = \$0\quad \forall j\in \mathcal{E}^{obPT} 
	\label{Eq:58}
\end{equation} 
\begin{equation}\begin{aligned}\label{Eq:59} 
	(v_{io}^\star x_{io} - \tau_o^\star)  \times q_{io}^\star &= \$0\quad \forall i \in I^C\\\quad [(v_{io}^\star -d_{io}^{x\star}) x_{io} - \tau_o^\star ]  \times q_{io}^\star &= \$0\quad \forall i \in I^O \\
    (u_{ro}^\star y_{ro} - \tau_o^\star)  \times p_{ro}^\star&= \$0\quad \forall r\in R^C\\\quad [(u_{ro}^\star-d_{ro}^{y\star}) y_{ro} - \tau_o^\star ] \times p_{ro}^\star&= \$0\quad \forall r\in R^O 
   \end{aligned} \end{equation}
The analysis of the osPT model is similar to that of the obPT model, as detailed in Section \ref{sec:4.6}, in which the performance metrics for the input and output are reversed. 

 In Equations (\ref{Eq:55})-(\ref{Eq:59}), one of the two terms on the left-hand side must become zero for their product to be equal to zero. SCSC fully supports the precise solutions of the Ord-VGA model. For instance, in Equation (\ref{Eq:56}), the linear combinations of the reference peers, in the set of $\mathcal{E}^{osPT}$, exceed the reduced value of output-r, $\relax [\sum_{\forall j\in \mathcal{E}^{osPT}} y_{rj}\pi_{oj}^\star - y_{ro}(1-p_{ro}^\star )>0]$ units and $u_{ro}^\star =\$0 $ per unit. On the other hand, $\relax [\sum_{\forall j\in \mathcal{E}^{osPT}} y_{rj}\pi_{oj}^\star - y_{ro}(1-p_{ro}^\star )=0]$ units and $u_{ro}^\star \geq\$0 $ per unit. 

 In Equation (\ref{Eq:57}), either $M_i^{xU}-(1+q_{io}^\star) x_{io}>0$ and $d_{io}^{x\star}=\$0$, or $M_i^{xU}-(1+q_{io}^\star) x_{io}=0$ and $d_{io}^{x\star}>\$0$.  In Equation (\ref{Eq:59}), $[(v_{io}^\star -d_{io}^{x\star}) x_{io} - \tau_o^\star ] < \$0$ and $q_{io}^\star=0$, otherwise, $[(v_{io}^\star -d_{io}^{x\star}) x_{io} - \tau_o^\star ]=\$0$  and $q_{io}^\star>0$.

When $(M_i^{xU} > x_{io})$ and the expression [$M_i^{xU}-(1+q_{io}^\star) x_{io}>0$], it necessitates and $d_{io}^{x\star}=\$0$. In Equation (\ref{Eq:45}), the term $ [(M_i^{xU}d_{io}^{x\star}+ x_{io}(v_{io}^\star-d_{io}^{x\star})]$ can be simplified to $v_{io}^\star x_{io}$. The adjusted $(1+q_{io}^\star) x_{io}>0$ might not yield an integer, but should range between two neighboring values on the Likert scale. The adjusted value is an approximation on the subjective Likert scale, and would not affect measuring the virtual gap $\Delta_o^{osPT\star}$. 

A special scenario occurs when the Likert scale of input-i matches or adjusts to the maximum, ($M_i^{xU}=x_{io}$) or  [$M_i^{xU}=(1+q_{io}^\star) x_{io}$]. In Equation (\ref{Eq:45}), the virtual price of input-i is tweaked by the penalty virtual price $-M_i^{xU}d_{io}^{x\star}$. A similar analysis holds for the situation where output-r has $M_r^{yL} =y_{ro}$ or adjusts to $M_r^{yL}=(1-p_{ro}^\star) y_{ro}$.

\subsection{Determining the Best DMU} \label{sec:5.5} In Stage I, the obPT model determines the reference group for $DMU_o$, designated as $\mathcal{E}^{obPT}_o$, which comprises DMUs that share input-output configurations similar to $DMU_o$. The set of leading DMUs compiled in $\mathcal{E}^{obPT}$ using Equation (\ref{Eq:25}) may not be uniform. During Stage II , the osPT model calculates a negative super-virtual gap for each $DMU_o$ in $\mathcal{E}^{obPT}$, denoted as $\delta_o^{osPT\star}(=\Delta_o^{osPT\star})$. This represents the degradation threshold when the inputs and outputs are scaled by the factors $Q_o^{\star}$ and $P_o^{\star}$. $DMU_{Best}$ surpasses its counterparts because it has the smallest negative super-virtual gap, as described in Equation (\ref{Eq:60}).
\begin{equation} \label{Eq:60}
    DMU_{Best} = \{ "Best" | \Delta_{Best}^{osPT\star}= \min_{\forall j \in \mathcal{E}^{obPT}} \Delta_j^{osPT\star} \}    
\end{equation}

\section{Explorations of the Derived Optimal Solutions}\label{sec:6} The subsequent sections discuss the primary advantages of scrutinizing the solutions derived from the obPT and osPT models. The Ord-VGA model possesses a \textit{unit invariance} characteristic, indicating that selecting different units of measurement (such as tons compared with kilograms) does not influence the evaluation outcomes. This feature facilitates a reliable assessment across various scales. 

\subsection{Calculating the Target Inputs and Outputs of Performance Metrics} \label{sec:6.1} 

This process involves the implementation of the obPT and osPT models, as described by Equations (\ref{Eq:61}) and (\ref{Eq:62}), respectively. This methodology demands that $DMU_o$ replicates its corresponding benchmark peers. This is achieved for each element $j$ within the sets $\mathcal{E}_o^{obPT}$ and $\mathcal{E}_o^{osPT}$ by utilizing the computed intensity values $\pi_{oj}^\star$. Here, for every $j\in \mathcal{E}_o^{obPT}$ and similarly for every $j\in \mathcal{E}_o^{osPT}$, these calculated intensities guide the duplication procedure, ensuring a detailed conformance to the established standards.
\begin{equation} \label{Eq:61}
\begin{aligned}
	&\widehat{x}_{io}^{obPT\star} = \sum_{\forall j\in \mathcal{E}_o^{obPT}} x_{ij}  \pi_{oj}^\star  = x_{io} (1 - q_{io}^\star )\quad \forall i\in I^C \cup I^O \\
	&\widehat{y}_{ro} ^{obPT\star} = \sum_{\forall j\in \mathcal{E}_o^{obPT}}y_{rj} \pi_{oj}^\star =y_{ro} (1 + p_{ro}^\star  )\quad  
 \forall r\in R^C \cup R^O
\end{aligned}
\end{equation}
\begin{equation}\label{Eq:62}
\begin{aligned}
	&\widehat{x}_{io}^{osPT\star} = \sum_{\forall j\in \mathcal{E}_o^{osPT}} x_{ij}  \pi_{oj}^\star  = x_{io} (1 + q_{io}^\star )\quad \forall i\in I^C \cup I^O \\
	&\widehat{y}_{ro} ^{osPT\star} = \sum_{\forall j\in \mathcal{E}_o^{osPT}}y_{rj} \pi_{oj}^\star =y_{ro} (1 - p_{ro}^\star  )\quad  
 \forall r\in R^C \cup R^O
\end{aligned}
\end{equation}

\subsection{Validate the Solutions}\label{sec:6.2}

The modified $X_i$ and $Y_r$ values of $DMU_o$ in the obPT and osPT models respectively, are outlined in Equations (\ref{Eq:61}) and (\ref{Eq:62}). It is essential that the following relationships hold:
\begin{equation} \begin{aligned} \label{Eq:63}
\widehat{x}_{io}^{obPT\star} \geq M_{io}^{xL}, \forall i \in I^O &\quad  \widehat{y}_{ro}^{obPT\star} \leq M_{ro}^{yU}, \forall r \in R^O \\
\widehat{x}_{io}^{osPT\star} \leq M_{io}^{xU},\forall i \in I^O &\quad \widehat{y}_{ro}^{osPT\star} \geq M_{ro}^{yL}, \forall r \in R^O.\\
\end{aligned}
\end{equation}
Utilizing Equations (\ref{Eq:61}) and (\ref{Eq:62}) for the obPT and osPT models, respectively, the \textit{benchmark virtual scales} (bvInput, bvOutput) are determined using Equation (\ref{Eq:64}). This process of deriving and fine-tuning the benchmarks illustrates how initial estimates from the two models are adjusted or calibrated, thereby narrowing the range of $\Delta_o^{obPT\star}$ and $\Delta_o^{osPT\star}$.
\begin{equation} \label{Eq:64} 
\begin{aligned}
	\widehat{\alpha}_o^b= \sum_{\forall i\in I^C \cup I^O} \widehat{x}_{io} v_{io}^\star, \quad &\widehat{\beta}_o^b= \sum_{\forall r\in R^C \cup R^O}\widehat{y}_{ro} u_{ro}^\star\\
    \widehat{\alpha}_o^s= \sum_{\forall i\in I^C \cup I^O} \widehat{x}_{io} v_{io}^\star, \quad &\widehat{\beta}_o^s= \sum_{\forall r\in R^C \cup R^O}\widehat{y}_{ro} u_{ro}^\star	
    \end{aligned}
	\end{equation}	
Furthermore, the points ($\widehat\alpha_o^b$, $\widehat\beta_o^b$) and ($\widehat\alpha_o^s$, $\widehat\beta_o^s$) correspond to a coordinate placed at the equator and prime meridian in the two-dimensional frameworks of the obPT and osPT solutions. Here, $\widehat\alpha_o^b=\widehat\beta_o^b$ and $\widehat\alpha_o^s=\widehat\beta_o^s$ are the essential conditions required to accurately assess the obPT and osPT models.

\subsection{Visualize the Virtual Technology Sets in 2D Graphics}\label{sec:6.3}  Equations (\ref{Eq:15}) and (\ref{Eq:16}) establish the virtual scale for each $DMU_j$ (\textit{vInput}, \textit{vOutput}). The virtual scales for each $DMU_j$ reflect particular aspects or features within the model. Equation (\ref{Eq:65}) defines the set of \textit{ virtual technologies} within the obPT model framework. For Equation (\ref{Eq:66}), associated with the osPT framework according to Equations (\ref{Eq:45}) and (\ref{Eq:46}), the following expression is presented:  
\begin{equation} \label{Eq:65}
    \Phi_o^{obPT\star}=\lbrace (\alpha_o^{b\star}, \beta_o^{b\star}) \mid (\alpha_j^{\star},\beta_j^{\star})\quad \forall j\in J\rbrace \quad \forall j \in J	
	\end{equation}
\begin{equation}	\label{Eq:66}
    \Phi_o^{osPT\star}=\lbrace (\alpha_o^{s\star}, \beta_o^{s\star}) \mid (\alpha_j^{\star},\beta_j^{\star})\quad \forall j\in J\rbrace \quad \forall j \in \mathcal{E}_o^{obPT}
	\end{equation}
The set of virtual technologies, denoted by $\Phi_o^{obPT\star}$, is associated with reference set $\mathcal{E}_o^{obPT}$. In the context of virtual gap analysis, it is presumed that MCA performance measures are divided into input metrics aimed at minimization and output metrics directed at maximization. The virtual technology set for $DMU_o$ can be graphically represented in two dimensions, as discussed in the following sections.

\subsection {Contrast the Advantages of Different Performance Metrics} \label{sec:6.4}
In the obPT and osPT models, the virtual cost ($\$$) for $DMU_o$ associated with each performance metric was derived using the formula provided in Equations (\ref{Eq:67}) and (\ref{Eq:68}).
\begin{equation} \label{Eq:67}
    \begin{aligned}
     &e_{io}^x=x_{io} v_{io}^\star\quad \forall i \in I^C,\\
     &e_{io}^{xL}=x_{io}( v_{io}^\star+d_{io}^{x\star})-M_{io}^{xL} \times d_{io}^{x\star}\quad \forall i \in I^O  \\
 &e_{ro}^y=y_{ro} u_{ro}^\star\quad \forall r \in R^C,\quad \\
 &e_{ro}^{yU}=y_{ro}(u_{ro}^\star+d_{ro}^{y\star})-M_{ro}^{yU}\times d_{ro}^{y\star}\quad \forall r \in R^O    \end{aligned}\end{equation}
 \vspace{-0.6 em}
\begin{equation} \label{Eq:68}
    \begin{aligned}
     &e_{io}^x=x_{io} v_{io}^\star\quad \forall i \in I^C,\\
     &e_{io}^{xU}=x_{io}( v_{io}^\star-d_{io}^{x\star})+M_{io}^{xU} \times d_{io}^{x\star}\quad \forall i \in I^O  \\
 &e_{ro}^y=y_{ro} u_{ro}^\star\quad \forall r \in R^C,\quad \\
 &e_{ro}^{yL}=y_{ro}(u_{ro}^\star-d_{ro}^{y\star})+M_{ro}^{yL}\times d_{ro}^{y\star}\quad \forall r \in R^O     
    \end{aligned}
\end{equation}
A performance metric with elevated virtual prices is more adept at pinpointing $DMU_o$'s best solutions. In Section \ref{sec:7.2}, solutions to the numerical example show that virtual prices for $DMU_K$ are represented by the values $(e_{1o}^x, e_{2o}^{xL}, e_{1o}^{yU}, e_{2o}^y)$, which are (0.6976, 0.3024, 0.3194, 0.3586). By dividing these virtual prices by the overall sum, which is $\$ $1.6972, we derive the normalized ratio as (0.4154: 0.1801: 0.1902: 0.2143). Moreover, one might separately examine the input and output ratios, ($e_{1o}^x: e_{2o}^{xL}$) and ($e_{1o}^{yU}: e_{2o}^y$) respectively. Furthermore, $DMU_o$ can leverage the estimated adjustment ratios ($ Q_o^\star, P_o^\star$) to effectively manage the performance metrics.

\section{Analyzing Numerical Case Studies} \label{sec:7} 

\subsection{The Minimal Working Examples}\label{sec:7.1}

Table \ref{table1} is a minimal illustrative example that has six decision-making units
(DMUs), designated as K, A, B, D, G, and H, employing four distinct performance metrics. This framework encompasses two input metrics, denoted as $X_1$ and $X_2$, which necessitate minimization, and  two output metrics, labeled as $Y_1$ and $Y_2$, which necessitate maximization. 

The evaluation of six unique laptop models (DMUs) utilizes quantitative performance metrics \(X_1\) and \(Y_2\) characterized by continuous, positive, and cardinal properties. Here, \(X_1\) represents the laptop's weight in kilograms per unit and \(Y_2\) refers to the number of units sold. On the other hand, qualitative metrics \(X_2\) and \(Y_1\) are expressed through Likert scales. Ordinal data \(X_2\) reflects brand perception on a scale from 1 ("very good") to 6 ("very bad"). Similarly, \(Y_1\) addresses user satisfaction and is divided into four levels from 4 ("very satisfied") to 1 ("very unsatisfied"). The Likert scale limits for \(X_2\) and \(Y_1\) were defined as \((1^{xL}, 6^{xL})\) and \((1^{yL}, 4^{yU})\), respectively.

\begin{table}[!ht]     
     \caption{An MCA example and decision variables associated to evaluate each Decision-Making Unit ($DMU_o$).} 
         \label{table1}
    \centering
     \setlength{\tabcolsep}{2 pt}
\renewcommand{\arraystretch}{1}
      \begin{tabular}{llccccccccccll} 
    
      \multicolumn{8}{c}{\textbf{ A minimal illustrative MCA example}} &\multicolumn{4}{c}{\quad} & \multicolumn{2}{c}{\textbf{Decision variables} }\\ \cline{1-8}
\multicolumn{2}{c}{Metrics (or Criteria)$\dagger$} &\multicolumn{6}{c}{Decision Matrix} &\multicolumn{4}{c}{\quad} & \multicolumn{2}{c}{of the metrics}\\  
\cline{1-2} \cline{13-14}\cline{9-10}
\multicolumn{2}{c}{\textit{m Inputs (unit)} \&}&\multicolumn{6}{c}{{\textit{n} DMUs (or Alternatives)}} & \multicolumn{4}{l}{\quad}  &Virtual price& Rate of \\ \cline{3-8}
\multicolumn{2}{c}{\textit{s Outputs (unit)}}  & K	&	A	&	B	&	D &	G &	H	 &\multicolumn{4}{c}{\quad} & per unit & adjust.\\ \cline{1-8} \cline{13-14}
$X_1$ Input:&$x_{1j}(kg)$   &1.6   &2.3   &1   &1.9   &1.8   &2.5   &\multicolumn{4}{c}{\quad}&$v_{1o}(\$/kg)$&$q_{1o}$ \\
$X_2$ Input:&$x_{2j}(pt.)$   &4   &3   &$6^{xU}$   &5  &3   &$1^{xL}$  &\multicolumn{4}{c}{\quad}&$(v_{2o}\pm d_{2o}^x)(\$/pt.)$&$q_{2o}$ \\
$Y_1$ Output: & $y_{1j}(lvl.)$ &  2   &3   & $1^{yL}$   &1   &2   & $4^{yU}$   &\multicolumn{4}{c}{\quad} & $(u_{1o}\pm d_{1o}^y)(\$/lvl.)$ & $p_{1o}$  \\
 $Y_2$ Output:&$y_{2j}(piece)$   &49   &97   &89   &97   &57   &70  &\multicolumn{4}{c}{\quad}&$u_{2o}(\$/piece$)&$p_{2o}$ \\
\cline{1-8} \cline{13-14}
\multicolumn{2}{c}{\textbf{Decision variables}}&$\pi_{oK}$&$\pi_{oA}$&$\pi_{oB}$ &
$\pi_{oD}$
&$\pi_{oG}$
&$\pi_{oH}$
&\multicolumn{4}{c}{\quad}& \textbf{Decision variables} & $\Delta_o^\star, \delta_o^\star$,\\
\cline{3-8} 
\multicolumn{2}{c}{of the DMUs} & \multicolumn{6}{c}{Intensity of each DMU-j } & \multicolumn{4}{c} {\quad} & of objective values &   $ \tau_o^\star \ddagger$  \\
\cline{1-8} \cline{13-14}
  \multicolumn{14}{l} {$\dagger$ Performance metrics of \textit{m} minimization criteria (inputs)}\\
  \multicolumn{14}{l} {$\qquad$ and \textit{s} maximization criteria (outputs).}  \\
  \multicolumn{14}{l}{$\ddagger$: $\Delta_o^\star$, the total virtual gap price; $ \delta_o^\star$, the total virtual adjustment price; $ \Delta_o^\star$=$\delta_o^\star$.}\\
   \multicolumn{14}{l}{ $\tau_o^\star$, an unified goal price of all metrics; $\$$, a virtual currency.}\\
 \end{tabular}
\end{table}

\subsection{Example Solutions from Table \ref{table1}} \label{sec:7.2} 
\begin{figure}[!ht]
\centerline{\includegraphics[width=4 in]{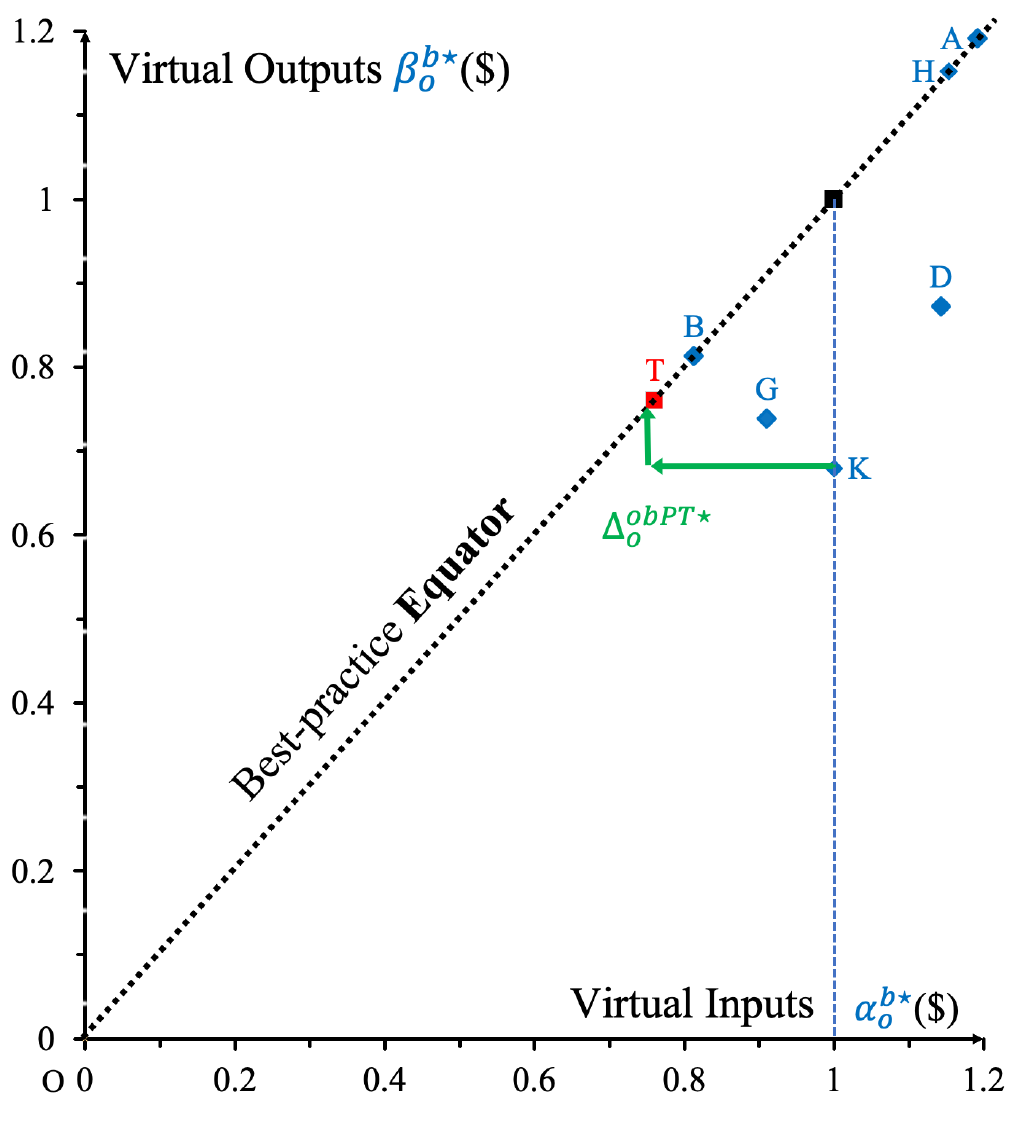}} 
\centering
 \caption {Assessment solutions of DMU-K in Stage I obPT model.} 
 \label{fig3}
  \end{figure}

Refer to the data presented in Table \ref{table2}. Figure \ref{fig3} visually supports the accuracy of the evaluation of $DMU_o$ and the $DMU_K$. The two-dimensional graph displays the six DMUs within the set $\Phi_K^{obPT\star}$. The solutions were analyzed with respect to their equator-aligned comparators $\mathcal{E}_K^{obPT}=\{A, B, H\}$. The diagonal is dubbed the equator because $DMU_K$ has its $\alpha_o^\star$ set at $\$1$ and is situated on the vertical axis at coordinates (1,0), extending north and south. The straight-line distance $\Delta_K^{obPT\star}$(=\$0.321) indicates the virtual price gap from point K to point T which is K's target value. Points G , D , and K were plotted below the equator.

\begin{figure}[!ht] \centerline{\includegraphics[width=4 in]{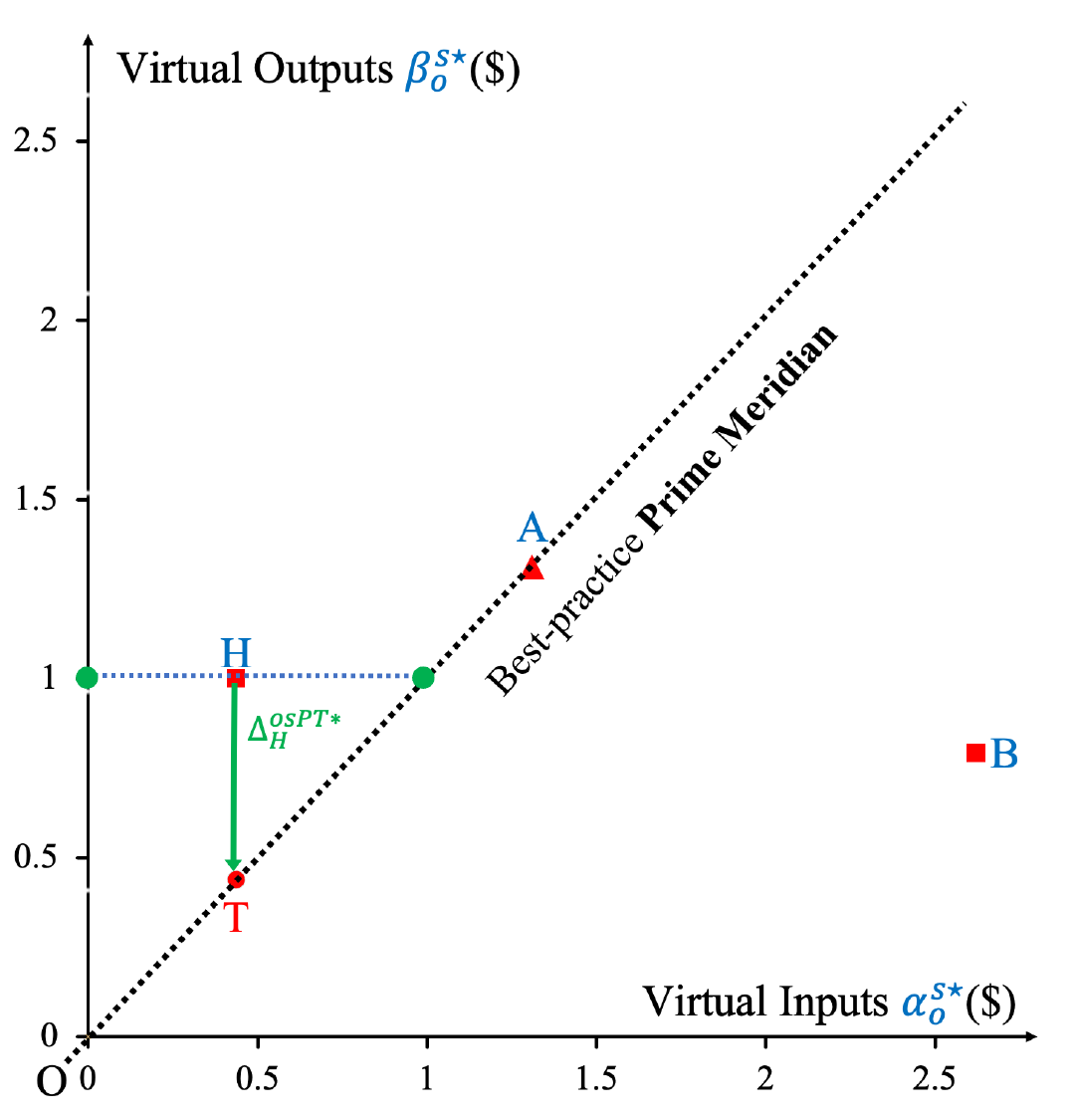}} 
 \caption {Assessment solutions of DMU-H in Stage II osPT model.} \label{fig4}
  \end{figure}

Figure \ref{fig4} illustrates the evaluation outcomes for $DMU_H$ during Stage II among the top three DMUs within the group $\Phi_H^{osPT\star}$. The R8 row records ($\Delta_{oA}^\star,\Delta_{oB}^\star, \Delta_{oH}^\star$) as (\$0, \$1.831, \$-0.563). Here, $\mathcal{E}_H^{osPT}=\{A\}$ with Point A located on the diagonal prime meridian. In contrast, Points H and B are positioned above and below the \textit{prime meridian}, with virtual deviations of -\$0.563 and \$0.183, respectively. The diagonal is designated as the prime meridian since $DMU_o$ has its $\beta_o^\star$ set at \$1 and is situated on the horizontal axis at (0, 1), extending east and west. $DMU_H$ reduces its second output $y_{2H}$ from 70 units to 32.3 units. The rectilinear distance $\Delta_H^{obPT\star}$ represents a \textbf{super-virtual gap} of -\$0.563, which spans from point H to point T and serves as the target for $DMU_H$.

  
\begin{table}[!htp]
\centering
\setlength{\tabcolsep}{1.2 pt}
\renewcommand{\arraystretch}{0.85}
\caption{Solutions in Stage I and Stage II Evaluations of DMUs in Table \ref{table1}.} \label{table2}
\begin{tabularx}{\linewidth}{lccrrrrrrcrrr}  \hline
&&Metric&\multicolumn{6}{c}{$DMU_o$ in Stage I }&&\multicolumn{3}{c}{$DMU_o$ in Stage II} \\ \cline{4-9} \cline{11-13}

Row&Symbol& unit & K	&	A	&	B	&	D &	G	& H	&&	A 	&	B 	&	$H\star$	\\  \cline{1-9} \cline{11-13} 
R1& $\tau_o^\star$	& $\$$	&	0.319	&	0.268	&	0.309	&	0.186	&	0.237	&	0.829	&&	1	&	1	&	0.437	\\
& $\Delta_o^\star, \delta_o^\star$ & $\$$ &	0.321	&	\textbf{0}	&	\textbf{0}	&	0.307	&	0.190	&	\textbf{0}	&&	\textbf{-0.103}	&	\textbf{-0.526}	&	\textbf{-0.563}	\\ \cline{1-9} \cline{11-13}
R2& $v_{1o}^\star$ & $\$$/kg &0.436	&	0.318	&	0.691	&	0.429	&	0.424	&	0.331	&&	0.244	&	0.474	&	0	\\
& $v_{2o}^\star$ & $\$$/(pt.) &	0.063	&	0.089	&	0.051	&	0.037	&	0.119	&	0.171	&&	0.112	&	0	&	0.437	\\
& $u_{1o}^\star$ & $\$$/(lvl.) & 0.160	&	0.089	&	0.309	&	0.186	&	0.119	&	0.043	&&	0	&	0	&	0.235	\\
& $u_{2o}^\star$ & $\$$/(piece)	&	0.007	&	0.008	&	0.008	&	0.005	&	0.010	&	0.012	&&	0.010	&	0.011	&	0.006	\\
& $d_{2o}^{x\star}$ & $\$$/(pt.)	&	0.02	&	0	&	0	&	0	&	0	&	0.657	&&	0	&	0	&	0	\\
& $d_{1o}^{y\star}$ & $\$$/(lvl.)	&	0	&	0	&	0	&	0	&	0	&	0.164	&&	0	&	0	&	0.126	\\ \cline{1-9} \cline{11-13}
R3& $q_{2o}^\star$ &-&	0.750	&	0	&	0	&	0.044	&	0.429	&	0	&&	0	&	0	&	0	\\
& $p_{1o}^\star$ &-&	0.254	&	0	&	0	&	1.609	&	0.371	&	0	&	& 0	&	0	&	0.750	\\
& $p_{2o}^\star$ &-&	0	&	0	&	0	&	0	&	0	&	0	&&	0.103	&	0.526	&	0.538	\\
& $\pi_{oA}^\star$ &-&	0.028	&	1	&	0	&	0	&	0.202	&	0	&&	0	&	\textbf{0.435}	&	0.333	\\
& $\pi_{oB}^\star$ &-&	0.054	&	0	&	1	&	0.718	&	0	&	0	&&	\textbf{0.371}	&	0	&	0	\\
& $\pi_{oH}^\star$ &-&	0.593	&	0	&	0	&	0.473	&	0.534	&	1	&&	\textbf{0.771}	&	0	&	0	\\ \cline{1-9} \cline{11-13}
R4& $v_{1o}^\star x_{1o}$ & $\$$	&	0.698	&	0.732	&	0.691	&	0.814	&	0.763	&	0.829	&&	0.561	&	0.474	&	0	\\
\multicolumn{3}{l}{($v_{2o}^\star \pm d_{2o}^{x\star}) x_{2o} \qquad \$$}	&	0.336	&	0.268	&	0.309	&	0.186	&	0.237	&	1.486	&&	0.337	&	0	&	0.437	\\
\multicolumn{3}{l}{($u_{1o}^\star \pm d_{1o}^{y\star}) y_{1o} \qquad \$$}	&	0.319	&	0.268	&	0.309	&	0.186	&	0.237	&	1.486	&&	0	&	0	&	0.311	\\ 
&$u_{2o}^\star y_{2o}$&$\$$&	0.360	&	0.732	&	0.691	&	0.508	&	0.573	&	0.829	&&	1	&	1	&	0.437	\\ \cline{1-9} \cline{11-13} 
R5& $\widehat\alpha_o^\star, \widehat\beta_o^\star$ & $\$$ &	0.760	&	1	&	1	&	0.992	&	0.898	&	1	&&	0.897	&	0.474	&	0.437	\\
& $\widehat{x}_{2o}^\star$ & (pt.) &	1	&	3	&	6	&	4.781	&	1.141	&	1	&&	3	&	6	&	1	\\
& $\widehat{y}_{1o}^\star$ & (lvl.)	&	2.509	&	3	&	1	&	2.609	&	2.742	&	4	&&	3	&	1	&	1	\\ 
& $\widehat{y}_{2o}^\star$ & (piece)	&	49	&	97	&	89	&	97	&	57	&	70	&&	87.1	&	55.6	&	\textbf{32.3}	\\ \cline{1-9} \cline{11-13}
R6& $\alpha_{oK}^\star$ & $\$$ &	1.000	&	0.866	&	1.312	&	0.834	&	1.153	&	1.216	&&		&		&		\\
& $\alpha_{oA}^\star$ & $\$$ &	1.191	&	1	&	1.744	&	1.097	&	1.331	&	1.277	&&	0.897	&	1.090	&	1.311	\\
& $\alpha_{oB}^\star$ & $\$$ &	0.813	&	0.854	&	1	&	0.651	&	1.136	&	1.360	&&	0.918	&	0.474	&	2.622	\\
& $\alpha_{oD}^\star$ & $\$$ &	1.143	&	1.051	&	1.570	&	1	&	1.398	&	1.487	&&		&		&		\\
&$ \alpha_{oG}^\star$ & $\$$ &	0.911	&	0.752	&	1.347	&	0.846	&	1.000	&	0.940	&&		&		&		\\
& $\alpha_{oH}^\star$ & $\$$ &	1.153	&	0.885	&	1.779	&	1.109	&	1.178	&	1	&&	0.722	&	1.185	&	0.437	\\ \cline{1-9} \cline{11-13}
R7& $\beta_{oK}^\star$ & $\$$ &	0.679	&	0.548	&	0.998	&	0.628	&	0.730	&	0.666	&&		&		&	\\
& $\beta_{oA}^\star$ & $\$$ &	1.191	&	1	&	1.680	&	1.064	&	1.331	&	1.277	&&	1	&	1.090	&	1.311	\\
& $\beta_{oB}^\star$ & $\$$ &	0.813	&	0.761	&	1	&	0.651	&	1.013	&	1.096	&&	0.918	&	1	&	0.791	\\
& $\beta_{oD}^\star$ & $\$$ &	0.872	&	0.822	&	1.062	&	0.693	&	1.093	&	1.191	&&		&		&		\\
& $\beta_{oG}^\star$ & $\$$ &	0.738	&	0.609	&	1.060	&	0.669	&	0.810	&	0.760	&&		&		&		\\
&$ \beta_{oH}^\star$ & $\$$ &	1.153	&	0.885	&	1.779	&	1.109	&	1.178	&	1	&&	0.722	&	0.787	&	1	\\ \cline{1-9} \cline{11-13}

R8& $\Delta_{oA}^\star$ &	$\$$&	0	&	0	&	\textbf{0.064}	&	0.033	&	0	&	0	&&	\textbf{-0.103}	&	0	&	0	\\
& $\Delta_{oB}^\star$ &	$\$$ &	0	&	\textbf{0.092}	&	0	&	0	&	0.123	&	\textbf{0.264}	&&	0	&	\textbf{-0.526}	&	1.831\\	
& $\Delta_{oH}^\star$ &	$\$$ &	0	&	0	&	0	&	0	&	0	&	0	&&	0	&	0.398	&	\textbf{-0.563$\star$}\\	 \hline
 \multicolumn{12}{l} {Note: All solutions of $q_{1o}^\star$, $\pi_{oK}^\star$,$\pi_{oD}^\star$, and $\pi_{oG}^\star$ are zero; all $\widehat{x}_{1o}^\star=x_{1o}$.}
 \end{tabularx}
 \end{table}

\subsection{A Unified Goal Price for each  \texorpdfstring{$DMU_o$}.}\label{sec:7.3}
In Table \ref{table2}, each $DMU_o$ defines a unified goal price, $\tau_o^\star$, which acts a boundary for the virtual pricing of metric inputs and outputs in the obPT and osPT models. Furthermore, metrics can exhibit both cardinal and ordinal data types, as shown in Equations (\ref{Eq:9}-\ref{Eq:12}) and (\ref{Eq:45}-\ref{Eq:48}), respectively. The model simultaneously optimizes objective values and both primal and dual decision variables, demonstrated in Table \ref{table2} as $\Delta_o^\star, (V_o^\star, U_o^\star, D_o^{x\star}, D_o^{y\star}),$ and $ (q_o^\star, p_o^\star, \Pi_o^\star)$.

\subsection{Handling of the Ordinal Data}\label{sec:7.4}
It is crucial to verify that treating ordinal data as continuous variables does not compromise the overall integrity of the analysis. In Table \ref{table1}, the qualitative indicators $X_2$ and $Y_1$, are depicted using Likert scales for the\textit{subjective evaluations}. Utilizing the solutions from column K of Table \ref{table2}, we computed the value of $x_{2o}$ using the SCSC equations (\ref{Eq:31}) and (\ref{Eq:7}):
\begin{equation*} 
\begin{aligned}
\quad [(v_{2o}^\star +d_{2o}^{x\star}) x_{2o} - \tau_o^\star ]=0\quad and \quad q_{2o}^\star >0 \\
(-M_2^{xL}+x_{2o}) \times d_{2o}^{x\star}= (-1+1)\times \$0.02=\$0. 
\end{aligned}
\end{equation*}

From Equation (\ref{Eq:31}), it follows that $[(u_{1o}^\star +d_{1o}^{y\star}) y_{1o} - \tau_o^\star ]=0$ while $p_{1o}^\star$ remains positive. The value of $y_{1o}$ increases from 2 to \textbf{2.509} (denoted as $\widehat{y}_{1o}^\star$), spanning Likert-scale values ranging from 2 to 3. As described in SCSC Equation (\ref{Eq:29}), the inequality $ [M_1^{yU}-(1+p_{1o}^\star) y_{ro}]>0$ holds true, with $d_{1o}^{y\star}$ equating to zero, and $\widehat{y}_{1o}^\star$ not reaching the Likert scale upper limit of 4. Additionally, based on Equation (\ref{Eq:7}), the expression $(-M_1^{yU}+y_{1o}) \times d_{1o}^{y\star}=0$ does not influence $\Delta_o^{obPT\star}$, including the term $-u_{1o}^\star \times y_{1o}$. The SCSC analysis corroborated the validity of the model.

The obPT and osPT models are inappropriate for handling \textbf{nominal data} metrics. For instance, in one metric, religious categories are numbered from one to five, rendering any continuous measurement within this metric pointless.

\subsection{Identification of Top Tier DMUs in Stage I}\label{sec:7.5} As shown in row R3 of Table \ref{table1}, the intensity values calculated in Stage I, represented as $\pi_{oA}^\star$, $\pi_{oB}^\star$, and $\pi_{oH}^\star$, and located in columns A, B, and H, respectively, all equate to 1, while the intensities elsewhere are zero. This indicates a clear divergence in the input-output configurations of the other DMUs compared to $DMU_o$.
Rows R6 and R7 enumerate the virtual input and output for each $DMU_j$ concerning $DMU_o$, identified by $\alpha_{oj}^\star$ and $\beta_{oj}^\star$. In the respective columns A, B, and H, the virtual gaps denoted by ($\alpha_{oA}^\star-\beta_{oA}^\star$, $\alpha_{oB}^\star-\beta_{oB}^\star$, $\alpha_{oH}^\star-\beta_{oH}^\star$)  are (0, 0.91, 0), (0.64, 0, 0), (0, 0.264, 0). This shows that $DMU_A$, $DMU_B$, and $DMU_H$ cannot be compared in Stage I.  

Section \ref{sec:6.2} outlines the validation process for the target $DMU_o$, which ensures that the virtual input $\widehat\alpha_o^\star$ matches the virtual output $\widehat\beta_o^\star$, as revealed in R5. Furthermore, referring to Equation (\ref{Eq:63}), the estimated ranking positions of performance metrics $X_2$ and $Y_1$ are indicated as $\widehat{x}_{2o}^\star$ and $\widehat{y}_{1o}^\star$ in row R5, illustrating the lower and upper limits on the Likert scales for Stages I and II.

\subsection{Compares the Top Tier DMUs in Stage II}\label{sec:7.6} As shown in Table \ref{table2}, during Stage II, the osPT model was used to assess the super-virtual gap for $DMU_A, DMU_B$, and $DMU_H$. The derived outputs, located in row R8 of Table \ref{table2}, were $-\$0.103$, $-\$0.526$, and $-\$0.563$, respectively. Considering the extent of the buffer degradation, the super-virtual gaps with the ranking order are $DMU_H>DMU_B>DMU_A$. According to Equation (\ref{Eq:65}), $DMU_H$ emerges as the ideal DMU for the MCA problem.

As shown in R3 row, decreasing $y_{1H}$ and $y_{2H}$ by 75\% and 53.8\% effectively eliminated the super-virtual gap for $DMU_H$, which is in line with the prime meridian. Figure \ref{fig4} demonstrates that $DMU_H$ is recognized as the top-performing DMU, denoted as $DMU_{Best}$. Position H is situated above the prime meridian with a significant virtual gap $\Delta_o^{osPT\star}$. Meanwhile, $DMU_A$ resides precisely on the prime meridian, featuring a zero virtual gap, indicated as $\pi_{oA}^{osPT\star}=0$, and has a reference set $\mathcal{E}_o^{osPT}$=\{A\}. The remaining peers are located beneath the prime meridian, each displaying a positive virtual gap.

$DMU_H$ benefits from the output metrics. Currently, the Likert scale rating for $y_{1H}$ peaks at 4, and can be lowered to $\widehat{y}_{1H}$, a minimum of 1, calculated as $4 \times (1 - 0.75)$. This was evaluated using Equation (\ref{Eq:38}), where $(M_1^{yL}-y_{1H})\times d_{1H}^{y\star}=(1-4)\times \$0.126= -\$0.378$. This calculation reveals the influence of the Likert scale on the super-virtual gap $\Delta_{oH}^{osPT\star}$.

\subsection{The real large size problem}\label{sec:7.7}
We appended the decision matrix \citep[digested from][Table 6]{Chen}, which was used to analyze energy efficiency in 29 Chinese provinces by employing three input and three output performance indicators, where the third indicator was evaluated using a Likert type scale. This case highlights the standard real-world MCA challenge, as shown in Tables \ref{table5} and \ref{table6}. 

Among the 29 DMUs, nine top-tier DMUs with high performance did not require modifications to inputs and outputs during Stage I evaluation, as indicated in rows R1 and R2 of Table \ref{table3}. 
\begin{table}[!ht]
\centering
\setlength{\tabcolsep}{3.2 pt}
\renewcommand{\arraystretch}{0.9}
\caption{Stage I Solutions of the obPT Ord-VGA models for each top-tier $DMU_o$ of the decision matrix \citep*[see][Table 5]{Chen}.} \label{table3}
 \footnotesize
\begin{tabularx}{\linewidth}{lccrrrrrrrrr}  \hline
Row&Sym-&Metric &\multicolumn{9}{c}{\textit{DMUo in top-tier subset}} \\ 
&	bol&	unit&	$DM_1$	&	$DM_4$	&	 $DM_5$	&	$DM_9$	&	$DM_{10}$	&	 $DM_{19}$	&	$DM_{21}$	&	$DM_{26}$	&	$DM_{28}$	\\ \hline
&&&\multicolumn{9}{l}{ $Q_o^\star=0, P_o^\star=0, \pi_o^\star=1, \pi_{oj}^\star=0 \quad \forall j \in \mathcal{E}^{obPT}$ }\\
R1& $E_o$	&-&	1	&	1	&	1	&	1	&	1	&	1	&	1	&	1	&	1	\\
& $\tau_o^\star$	& $\$$	&	0.0283	& 0.0110	& 0.0541 &	0.0788 &	0.0115 &	0.0218 &	0.2470	& 0.0066	& 0.3028	\\
& $\Delta_o^\star, \delta_o^\star$ & $\$$ &	0	&	0	&	0	&	0	&	0	&	0	&	0	&	0	&	0	\\ \hline
R2& $v_{1o}^\star$ & $\$$/Rm &	1E-04	&	1E-04	&	1E-04	&	1E-04	&	1E-04	&	1E-04	&	1E-04	&	1E-04	&	1E-04	\\
& $v_{2o}^\star$ & $\$$/Pn &	3E-08	&	3E-09	&	1E-07	&	1E-08	&	6E-08	&	5E-08	&	3E-07	&	2E-09	&	4E-07	\\
& $v_{3o}^\star$ & $\$$/Tn &	6E-05	&	3E-05	&	3E-06	&	7E-06	&	4E-07	&	7E-07	&	3E-04	&	9E-05	&	7E-05	\\
& $u_{1o}^\star$ & $\$$/Rm &	5E-05	&	2E-05	&	5E-05	&	4E-05	&	2E-05	&	2E-05	&	2E-04	&	2E-05	&	1E-04	\\
& $u_{2o}^\star$ & $\$$/Tn&	3E-06	&	1E-05	&	3E-06	&	9E-06	&	7E-07	&	5E-07	&	7E-05	&	2E-05	&	3E-05	\\
& $u_{3o}^\star$ & $\$$/lvl. &	3E-02	&	5E-03	&	3E-02	&	8E-02	&	1E-02	&	1E-02	&	7E-02	&	7E-03	&	2E-01	\\ 
& $d_{3o}^{y\star}$ & $\$$/lvl. &0	&0	&0	&0	&0	&0	&1.6E-02	&0	&0	\\ \hline
R3& $v_{1o}^\star x_{1o}$ & $\$$ 	&	0.325	&	0.354	&	0.551	&	0.842	&	0.528	&	0.298	&	0.283	&	0.007	&	0.394	\\
& $v_{2o}^\star x_{2o}$ & $\$$	&	0.240	&	0.011	&	0.395	&	0.079	&	0.460	&	0.680	&	0.247	&	0.007	&	0.303	\\
& $v_{3o}^\star x_{3o}$ & $\$$	&	0.435	&	0.636	&	0.054	&	0.079	&	0.012	&	0.022	&	0.470	&	0.987	&	0.303	\\
& $u_{1o}^\star y_{1o}$ & $\$$	&	0.943	&	0.222	&	0.742	&	0.753	&	0.947	&	0.956	&	0.555	&	0.296	&	0.303	\\
& $u_{2o}^\star y_{2o}$ & \$	&	0.028	&	0.767	&	0.204	&	0.168	&	0.042	&	0.022	&	0.247	&	0.698	&	0.394	\\
\multicolumn{3}{c}{$(u_{3o}^\star+d_{3o}^{y\star}) y_{3o} \$$}	&	0.028	&	0.011	&	0.054	&	0.079	&	0.012	&	0.022	&	0.247	&	0.007	&	0.303	\\ \hline
\multicolumn{11}{l}{Note:
Rm= (RMB), Pn=(person), Tn(Tons), lvl.=(level of Likert scale)}.
 \end{tabularx}
 \end{table}
In Row R3, the virtual prices for the input and output metrics are not less than $\tau_o^\star$. For further details, see Equation (\ref{Eq:9})-Equation (\ref{Eq:12}).
\begin{figure}[!ht] \centerline{\includegraphics[width=4 in]{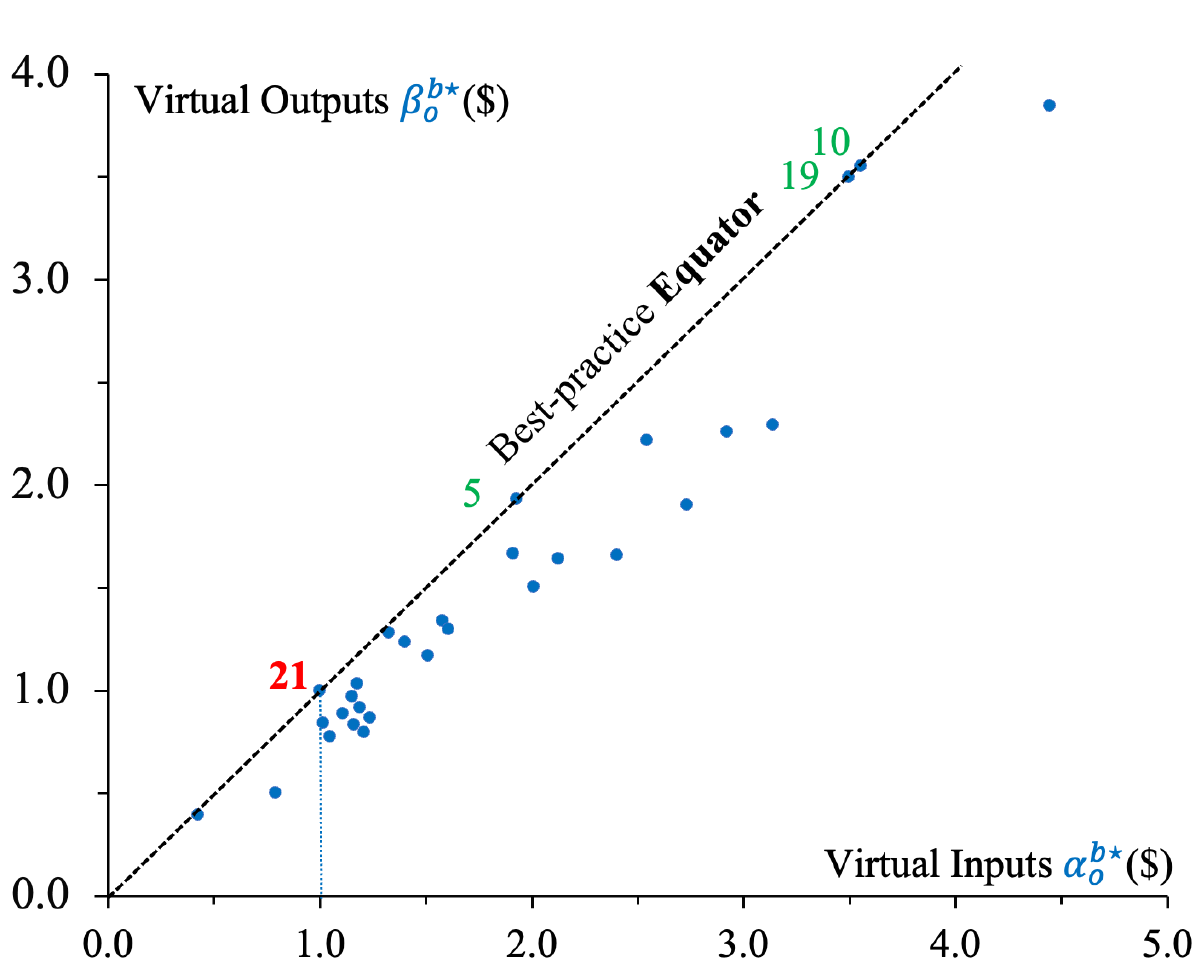}} 
 \caption {Assessment solutions of DMU-21 in Stage I.} \label{fig5}
  \end{figure}

 Figure \ref{fig5} illustrates the evaluation outcomes for $DMU_{21}$ in Stage I. For this particular unit, only $\pi_{21}^{obPT\star}$ holds a value of 1, while all other $DMUj$ display $\pi_j^{obPT\star}$ as 0. The DMUs numbered 21, 5, 19, and 10 all exhibited virtual gaps of zero and were positioned at the equator. By contrast, the other 25 DMUs were situated beneath the equator and had positive virtual gaps.
For the 29 DMUs considered in this instance, the solutions for assessing $DMU_o$ are illustrated in the corresponding graph. In this setting, $DMU_o$ must be located on the vertical axis at coordinates (1, 0), allowing it to move vertically. 

\begin{table}[!htp]
\centering
\setlength{\tabcolsep}{0.55 pt}
\renewcommand{\arraystretch}{0.8}
\caption{Stage II Solutions of the osPT Ord-VGA models for each $DMU_o$ of \citep{Chen}.} \label{table4}
 %\footnotesize
\begin{tabularx}{\linewidth}{lccrrrrrrrrr}  \hline
Row&Sym-&Metric &\multicolumn{9}{c}{\textit{DMUo in the top-tier subset}} \\ 
&	bol&	unit&	$DM_1$	&	$DM_4$	&	 $DM_5$	&	$DM_9$	&	$DM_{10}$	&	 $DM_{19}$	&	$DM_{21}$	&	$DM_{26}$	&	$DM_{28}$\\ \hline
R1& $\tau_o^\star$	& $\$$	&	0.991	&	0.800	&	0.547	&	0.759	&	1	&	1	&	0.363	&	0.926	&	0.779	\\
& $\Delta_o^\star, \delta_o^\star$ & $\$$&	-0.216	&	-0.036	&	-0.235	&	-0.254	&	-0.271	&	-0.066	&	-0.496	&	-0.048	&	-0.231	\\ \hline
R2& $v_{1o}^\star$ & $\$$/Rm &	0	&	9E-05	&	0	&	1E-04	&	0	&	9E-06	&	0	&	0	&	3E-04	\\
& $v_{2o}^\star$ & $\$$/Pn &	0	&	0	&	2E-07	&	0	&	7E-08	&	2E-08	&	2E-07	&	6E-09	&	3E-07	\\
& $v_{3o}^\star$ & $\$$/Tn &	1E-04	&	8E-06	&	1E-05	&	0	&	5E-06	&	2E-05	&	2E-04	&	9E-05	&	0	\\
& $u_{1o}^\star$ & $\$$/Tn &	6E-05	&	2E-05	&	3E-05	&	4E-05	&	2E-05	&	2E-05	&	1E-04	&	3E-05	&	0	\\
& $u_{2o}^\star$ & $\$$/Rm&	0	&	1E-05	&	8E-06	&	1E-05	&	0	&	0	&	0	&	2E-05	&	5E-05	\\
& $u_{3o}^\star$ & $\$$/lvl. &	9E-03	&	0	&	0	&	1E-02	&	0	&	0	&	4E-01	&	0	&	2E-01	\\
& $d_{3o}^{y\star}$ & $\$$/lvl.	&	0	&	0	&	0	&	0	&	0	&	0	&	\textbf{0.274}	&	0	&	0	\\ \hline
R3& $v_{1o}^\star x_{1o}$ & $\$$ &		0	&	0.800	&	0	&	0.746	&	0	&	0.160	&	0	&	0	&	0.553	\\
& $v_{2o}^\star x_{2o}$ & $\$$	&		0	&	0	&	0.547	&	0	&	0.572	&	0.232	&	0.141	&	0.026	&	0.216	\\
& $v_{3o}^\star x_{3o}$ & $\$$	&		0.784	&	0.164	&	0.219	&	0	&	0.157	&	0.541	&	0.363	&	0.926	&	0	\\
& $u_{1o}^\star y_{1o}$ & $\$$	&		0.991	&	0.200	&	0.453	&	0.759	&	1.000	&	1.000	&	0.363	&	0.397	&	0	\\
& $u_{2o}^\star y_{2o}$ & $\$$	&		0	&	0.800	&	0.547	&	0.228	&	0	&	0	&	0	&	0.603	&	0.779	\\
\multicolumn{3}{c}{$(u_{3o}^\star-d_{3o}^{y\star}) y_{3o} \$$}		&		0	&	0	&	0	&	0.013	&	0	&	0	&	0.363	&	0	&	0.442	\\ \hline
R4& $q_{1o}^\star$	&	-&	0	&	0.001	&	0	&	0	&	0	&	0	&	0	&	0	&	0	\\
& $q_{2o}^\star$	&	-&	0	&	0	&	0.298	&	0	&	0	&	0	&	0	&	0	&	0	\\
 &$q_{3o}^\star$	&	-&	0	&	0	&	0	&	0	&	0	&	0	&	0.591	&	0.052	&	0	\\
 &$p_{1o}^\star$	&	-&	0.218	&	0	&	0	&	0.334	&	0.271	&	0.066	&	0.108	&	0	&	0	\\
& $p_{2o}^\star$	&	-&	0	&	0.045	&	0.131	&	0	&	0	&	0	&	0	&	0	&	0.297	\\
 &$p_{3o}^\star$	&	-&	0	&	0	&	0	&	0	&	0	&	0	&	0.667	&	0	&	0	\\ \hline
R5& $\pi_{o1}^\star$ &-	&	0	&	0	&	0	&	0	&	0	&	0.133	&	0.082	&	0.404	&	0	\\
& $\pi_{o4}^\star$ &-	&	0	&	0	&	0	&	0.089	&	0	&	0	&	0	&	0	&	0.042	\\
& $\pi_{o5}^\star$ &-	&	0	&	0	&	0	&	0	&	0.748	&	0	&	0	&	0.391	&	0.102	\\
& $\pi_{o9}^\star$ &-	&	0	&	0.153	&	0	&	0	&	0	&	1.736	&	0	&	0	&	0	\\
& $\pi_{o10}^\star$ &-	&	0	&	0	&	0.146	&	0	&	0	&	0.293	&	0	&	0.019	&	0	\\
& $\pi_{o19}^\star$ &-	&	0.236	&	0	&	0	&	0.208	&	0.482	&	0	&	0	&	0	&	0	\\
& $\pi_{o21}^\star$ &-	&	0.176	&	0	&	0	&	0	&	0	&	0	&	0	&	0	&	0.237	\\
& $\pi_{o26}^\star$ &-	&	0	&	0	&	0	&	0	&	0	&	0	&	0	&	0	&	0	\\
& $\pi_{o28}^\star$ &-	&	0	&	3.858	&	3.414	&	0.203	&	0	&	0	&	0.459	&	0	&	0	\\ \hline
R6& $\alpha_o^\star$ &$\$$	&	0.784	&	0.964	&	0.765	&	0.746	&	0.729	&	0.934	&	0.504	&	0.952	&	0.769	\\
& $\beta_o^\star$ &$\$$	&	1	&	1	&	1	&	1	&	1	&	1	&	1	&	1	&	1	\\
& $\widehat{\alpha}_o^\star, \widehat{\beta}_o^\star$ &$\$$	&	0.784	&	0.964	&	0.928	&	0.746	&	0.729	&	0.934	&	0.719	&	1	&	0.769	\\ \hline
R7& $\Delta_{o1}^\star$ &	$\$$ &	\textbf{-0.216}	&	0.198	&	0.942	&	0.085	&	0.202	&	0	&	0	&	0	&	3.201	\\
& $\Delta_{o4}^\star$ &	$\$$ &	1.422	&	\textbf{-0.036}	&	0.242	&	0	&	0.181	&	0.300	&	2.513	&	0.172	&	0	\\
& $\Delta_{o5}^\star$ &	$\$$ &	1.262	&	0.113	&	\textbf{-0.235}	&	0.232	&	0	&	0.239	&	1.872	&	0	&	0	\\
& $\Delta_{o9}^\star$ &	$\$$ &	0.113	&	0	&	0.529	&	\textbf{-0.254}	&	0.071	&	0	&	0.354	&	0.132	&	1.985	\\
& $\Delta_{o10}^\star$ &	$\$$ &	0.143	&	1.419	&	0	&	1.725	&	\textbf{-0.271}	&	0	&	0.240	&	0	&	7.629	\\
& $\Delta_{o19}^\star$ &	$\$$ &	0	&	0.443	&	0.976	&	0	&	0	&	\textbf{-0.066}	&	0.267	&	0.219	&	6.404	\\
& $\Delta_{o21}^\star$ &	$\$$ &	0	&	0.114	&	0.090	&	0.120	&	0.018	&	0.016	&	\textbf{-0.496}$\star$	&	0.004	&	0\\	
& $\Delta_{o26}^\star$ &	$\$$ &	0.350	&	0.539	&	0.283	&	0.795	&	0.073	&	0.120	&	0.698	&	\textbf{-0.048}	&	2.630\\	
& $\Delta_{o28}^\star$ &	$\$$ &	0.351	&	0	&	0	&	0	&	0.028	&	0.074	&	0	&	0.052	&	\textbf{-0.452}	\\ \hline
\multicolumn{12}{l}{Note: Rm= (RMB), Pn=(person), Tn(= tons), lvl.=(levels on the Likert scale).}
 \end{tabularx}
 \end{table}

Outlined in Table \ref{table4}, the Stage II evaluation of the nine leading DMUs of the decision matrix is described by \citet{Chen}. Row R3 indicates that virtual prices remain below or equal to $\tau_o^\star$, following the criteria established in Equation (\ref{Eq:40}) through Equation (\ref{Eq:43}).

In row R3, according to Equation (\ref{Eq:60}), $(u_{3,21}^\star-d_{3,21}^{y\star}) y_{3,21}=\tau_{21}^\star=\$0.363$, hence $q_{o,21}^\star>0$. The expression in Equation (\ref{Eq:38}) is $(M_3^{yL}-3)\times d_{3,21}^{y\star}=(1-3)\times \$0.274= -\$0.548$. This quantifies the impact of the Likert scale on the super-virtual gap $\Delta_{21}^{osPT\star}$. 

Row R4 showed an increase of 59.1\% in $x_{3, 21}$, along with decreases of 10.8\% and 66.7\% in $y_{1, 21}$ and $y_{3, 21}$, respectively, leading to a virtual zero gap, aligning it with the prime meridian. Currently, $y_{3, 21}$ is the ceiling Likert scale of 3 and will be reduced to the floor Likert scale of 1, calculated as $y_{3, 21} \times (1 - p_{3,21}^\star)=1$. 

\begin{figure}[!ht] \centerline{\includegraphics[width=3.7 in]{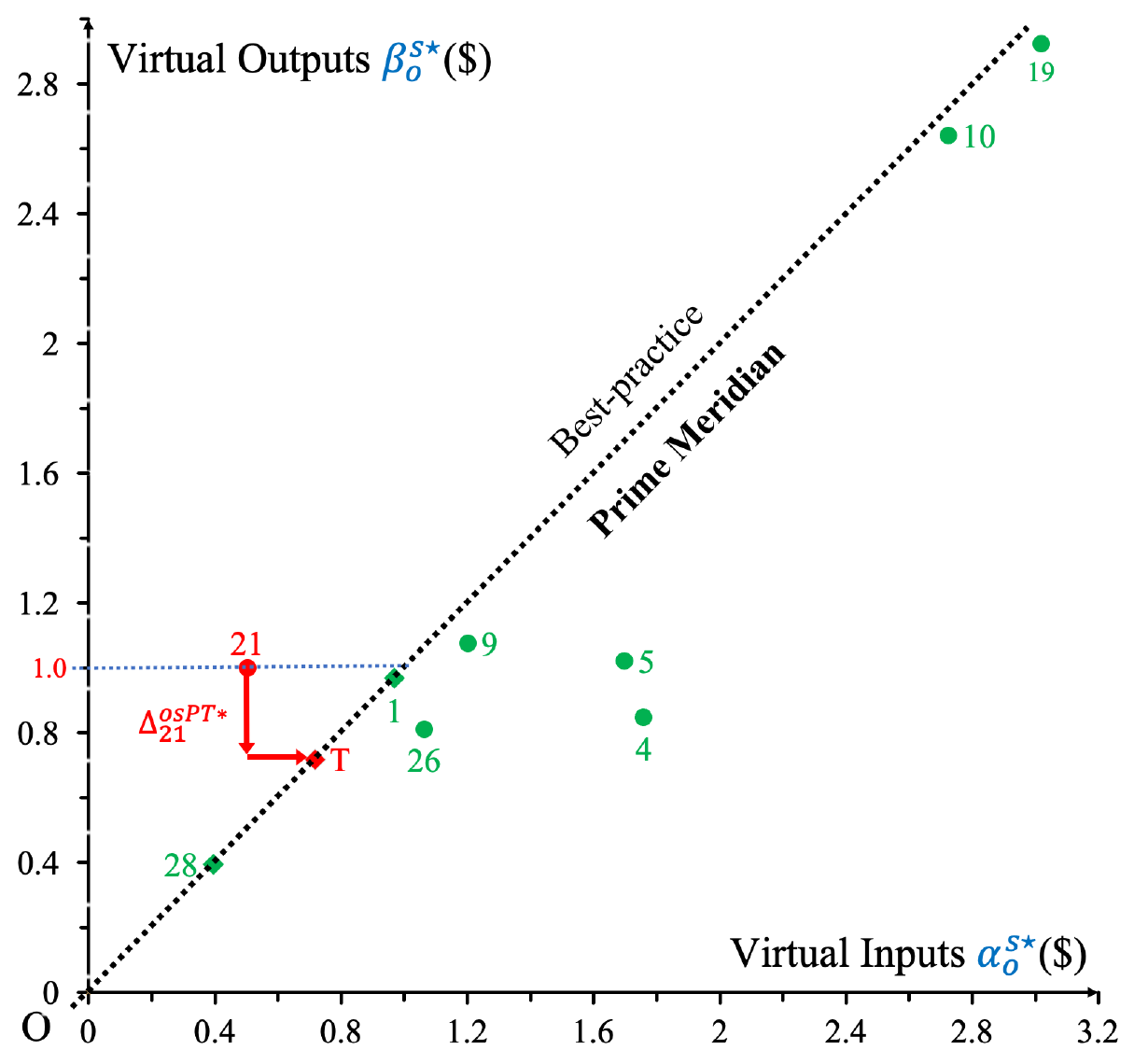}} 
 \caption {Assessment solutions of DMU-21 in Stage II.} \label{fig6}
  \end{figure}
  
  Figure \ref{fig6} illustrates the methods used to evaluate the super-virtual gap of $DMU_{21}$ in Stage II. The coordinates of the positions for points 21 and T are listed as (\$0.504, \$1) and (\$0.719, \$0.719), respectively, as shown in row R6. The rectilinear distance from point 21 to point T was equal to $\Delta_{o,21}^{osPT\star}=\$ -0.496$. Reference peers $DMU_1$ and $DMU_{28}$ and T were located on the prime meridian. The other six DMUs were located below the prime meridian.

A comparable graph illustrating the resolution for evaluating the super-virtual gap of each $DMUo$ was obtained. Applying Equation (\ref{Eq:60}), Hainan Province $DMU_{21}$ emerges as the foremost DMU in the MCA framework, presenting a super-virtual gap of -\$0.496. Refer to R7 for further details.

Within the linear programming frameworks of the obPT and osPT models, the Likert scales for the performance metrics were treated as continuous values, thereby optimizing the computational efficiency. These models effectively evaluate the virtual price corresponding to each Likert scale level, specifically, $D_o^x$ and $D_o^y$. The decision matrix of Table \ref{table6} with 6 rows and 29 columns \citep{Chen} was solved within minutes.

\section{Discussions and Conclusions} \label{sec:8} 
The innovative multiple-criteria assessment methodology proposed herein, which addresses the heterogeneity among DMUs and utilizes ordinal data, is intended for practical applications. The approaches outlined to tackle the MCA challenge may be compared with the existing DEA, SFA, and MCDM methodologies. The considerations articulated in the DEA, SFA, and MCDM models provide valuable insights that can be incorporated into the Ord-VGA models.

\subsection{Heterogeneity DMUs} \label{sec:8.1} Typically, traditional MCA approaches assume that DMUs (Decision Making Units) function under a consistent input-output structure, evaluating them through a set of \textbf{weights} for performance metrics expressed in cardinal data with varying measurement units. Several studies \cite{Aleskerov, Zhu, 2016LiW, Zarrin}, have raised concerns that heterogeneous DMUs may lead to unreliable evaluations. Our findings indicate that this heterogeneity is related to the measurement units applied to performance metrics. Establishing guidelines for selecting suitable measurement units and precisely defining what constitutes heterogeneity remains a challeng.

 In the current research, the MCA decision matrices contain cardinal and ordinal data, expressed in real values and Likert scales. The MCA is applicable to daily life in various situations. Instead of choosing measurement units and weights of the performance metrics, the proposed MCA presents a significant advantage over the traditional MCA methodologies. Sections \ref{sec:4.3} and \ref{sec:5.3} elucidated the processes for deriving the \textbf{unified goal price of $DMU_o$}, $\tau_o^\star$ for the obPT and osPT models. 
 
  In Step 1 of the obPT primal program, Equations (\ref{Eq:9}) through (\ref{Eq:12}) constrain each \textbf{virtual price} to the lower bound \textbf{unified goal price of $DMU_o$}, $\tau_o^\# (=\$1)$. In regardless of the measurement units used to the performance metrics, which are measured in a virtual currency $\$$ in the obPT model. According to Equations (\ref{Eq:27}), (\ref{Eq:28}) and (\ref{Eq:30}), $DMU_j$ is identified as a reference peer of $DMU_o$, and has a zero virtual gap and $\pi_{oj}^\#>0$. The obPT model simultaneously estimated the decision variables $(V_o^\#, U_o^\#, D_o^{x\#}, D_o^{y\#})$ and $(Q_o^\#, P_o^\#, \Pi_o^\#)$. The model determines the reference set  $\mathcal{E}_o^{obPT}$,  to achieve the optimal objective values. Utilizing the obPT model does not involve the issue of homogeneous or heterogeneous input-output frameworks of DMUs. Equation (\ref{Eq:24}) defines the connection between the solutions of Steps I and II.  
 
 The interpretations concerning the osPT model are analogous and were therefore omitted. 

\subsection{The Treatments of Ordinal Data}\label{sec:8.2}
In practical applications, decision matrices inherently comprise of cardinal and ordinal data. Within the Ord-VGA models, the bounded integral values of ordinal data are treated as continuous real values with specified bounds. If the estimated Likert scale for a metric of ordinal data reaches a boundary, a penalty price should be imposed. Conversely, if the estimated Likert scale remains within the established bounds, the virtual price per unit of the metric remains unaffected. The strong complementary slackness conditions (SCSC) detailed in Sections \ref{sec:4.6} and \ref{sec:5.4} serve to validate the two Ord-VGA models.

\subsection{Approaches to Address the MCA Challenge} \label{sec:8.3}
Various MCA techniques have been formulated to address diverse decision-making problems, each with distinct advantages and limitations. To ensure clarity and methodological accuracy, we detail the key approaches for addressing the MCA challenges.

In Sections \ref{sec:4} and \ref{sec:5}, a thorough explanation of the two Ord-VGA models is presented. MCA issues require consideration of both the minimization and maximization parameters, which are referred to as input and output metrics, respectively.
\begin{enumerate}
\item {Without Assumptions}: The two Ord-VGA models were established without assumptions. 
\item {Criteria Conversion}: Various measurement units are transformed into a \textit{virtual price} for each criterion (or performance metric) which involves multiplying the value by its virtual unit price, expressed in virtual currency $\$$.
\item {Multiple Criteria Integration}: The virtual prices for both input and output metrics are amalgamated into what is termed the \textit{virtual input price} and \textit{virtual output price}. 
\item {Unified goal price}: In evaluating $DMU_o$ through each Ord-VGA model, the two-step method ensures that the virtual gap of $DMU_o$ resides in the interval between ($\$0$, $\$1$).
\item {Addressing Diverse DMUs}:
Typically, many MCA approaches demand \textit{uniform} input-output structures in decision matrices, which exposes them to difficulties with normalization and weighting. In contrast, the novel method effectively handles heterogeneous DMUs.
\item {Dealing with Likert Scale Metrics}: When managing Likert scale metrics with defined maximum and minimum values, the metric for $DMU_o$ involves a \textit{modified virtual unit price}. 
\item {Removal of Subjective Assessments}: Current MCA methods are frequently plagued by subjective evaluations and biases, which undermine the reliability of the results despite significant efforts.
\item {Two-stage Ord-VGA Evaluation}: Initially, the primary stage pinpoints the top DMUs from among the \textit{n} DMUs. In the second stage, the leading DMU is selected among the top DMUs.  
\item {Conditions of Duality}: The validity of duality is affirmed through the analysis of the strong complementary slackness conditions (SCSC) that connect the \textit{primal and dual programs} in each Ord-VGA model. The development of the obPT and osPT models has been well demonstrated.
\item {Assessing the virtual gaps of $DMU_o$ across the two stages} the virtual gap between the virtual input and output prices in both stages was examined.
\item {Graphical depiction}: Each evaluation outcome for $DMU_o$ is illustrated using \textit{2D graphical intuition} to verify accuracy of the methodology. The diagonal alignment of the equator and prime meridian defines the perimeter of the virtual technology set during Stages I and II.
\item {Ensuring comprehensive assessments}: The updated $DMU_o$ in both stages should not display a virtual gap and must lie on the equator and prime meridian. This indicated that the two models provided detailed evaluations.
\item {Scalability and Computational Performance}: Finally, the speed of computation and the greatest feasible problem size depend on the effectiveness of the linear programming software utilized.
\item {Ranking DMUs}: Remove the top-performing DMU from the decision matrix and repeat the evaluation process to determine the following best DMU.
\end{enumerate}

\subsection{Research Opportunities} \label{sec:8.4} 
Literature concerning traditional multi-criteria analysis (MCA) methods serves as a foundation for enhancing the applicability of assessments. 
\begin{enumerate}
\item {Identifying the alternative with the poorest performance is equally important. A forthcoming article tackles the issue of worst practices in the MCA.} 
\item {VGA models do not permit decision matrices that contain zero or negative values, thereby presenting a rich area for exploration.} 
\item {It would be advantageous to integrate the various side constraints explored in the DEA, SFA, and MCDM methodologies into Ord-VGA models.}
\end{enumerate}

\subsection*{Acknowledgments}
This study was inspired by the DEA and MCDM methods. The authors thank the referees for providing their valuable feedback. 

\subsection*{Declaration of competing interest}
This research did not receive specific grants from public, commercial, or nonprofit funding agencies. The authors have no competing interests to disclose. 
\subsection*{CRediT authorship contribution statement}

\textbf{Fuh-Hwa Franklin Liu}: Conceptualization, Methodology, Writing of the Original Draft, Supervision. 
\textbf{Su-Chuan Shih}: Conceptualization, Data collection, Validation, Formal Analysis, Visualization, Review and Edition.

\subsection*{Data availability}
We confirm that data supporting the findings of this study are available in this article.

\medskip
\providecommand{\url}[1]{\texttt{#1}}
\providecommand{\urlprefix}{URL }
\section*{Appendix} Tables \ref{table5} and \ref{table6} are derived from \cite{Chen}. The data analysis employs three distinct models to calibrate the outputs \citep[see][Models (2), (3), and (4)]{Chen}. Subsequently, three solutions were proposed \citep[see][Tables 6, 7, and 8]{Chen}. Table \ref{table4} shows our assessments of the nine foremost DMUs, wherein all inputs and outputs were systematically adjusted. The DMU-21 surpassed the others.
\begin{table}[!ht]
\setlength{\tabcolsep}{2 pt}
 \caption{Input and output measures for energy efficiency application. \citep[Table 4 of][]{Chen}}
    \centering \label{table5}
    \begin{tabular}{llllr}
        \hline
        \textbf{Measure} & \textbf{Unit} & \textbf{Type}&\textbf{Data} &Metric\\
        \hline
        Capital (total investments  & 100 million RMB & Real &Cardinal&X1\\
    \qquad \qquad in fixed assets) \\
        Labor & Person & Integer&Cardinal &X2\\
        Energy (total energy & 10,000 tons of & Real &Cardinal&X3\\
       \qquad \qquad consumption)& \qquad coal equivalent \\
      
        GDP (gross domestic product) & 100 million RMB & Real&Cardinal &Y1\\
        CO2 emissions & 10,000 tons & Real &Cardinal&Y2\\
        Air quality level & None & Likert scale &Ordinal&Y3\\
        \hline
    \end{tabular}
\end{table}
 \begin{table}[!ht]
\centering
\setlength{\tabcolsep}{4 pt}
\renewcommand{\arraystretch}{0.9}
\caption{Decision matrix of 29 Provinces. \citep[Table 5 of][]{Chen}.} \label{table6}
 \footnotesize
\begin{tabularx}{\linewidth}{lrrrrrrr}  \hline
&& \multicolumn{6}{c}{Measure (Metric)}\\ \cline{3-8}
Province	&	DMU	&	Capital	&	Labor	&	Energy	&	GDP	&	CO$_2$	&	Air\\ 
	&		&	X1 	&	X2 	&	X3 	&	Y1 	&	Y2 &	Y3 	\\ \hline
Beijing	&	DM-1	&	6112.3714	&	7173717	&	7177.6819	&	17879.4	&	9582.5296	&	1	\\
Tianjin	&	DM-2	&	7934.7828	&	2890714	&	8208.0083	&	12893.88	&	12523.8754	&	1	\\
Hebei	&	DM-3	&	19661.2832	&	6199464	&	30250.2137	&	26575.01	&	60245.3607	&	2	\\
Shanxi	&	DM-4	&	8863.2642	&	4360001	&	19335.5884	&	12112.83	&	64400.8714	&	2	\\
Inner Mongolia	&	DM-5	&	11875.7409	&	2707695	&	19785.7079	&	15880.5788	&	69575.0256	&	2	\\
Liaoning	&	DM-6	&	21836.2831	&	5987272	&	23526.4048	&	24846.43	&	41316.6819	&	2	\\
Jilin	&	DM-7	&	9511.5371	&	2854781	&	9443.0384	&	11939.24	&	22145.1092	&	2	\\
Heilongjiang	&	DM-8	&	9694.7461	&	4709834	&	12757.8001	&	13691.58	&	29326.8357	&	2	\\
Shanghai	&	DM-9	&	5117.6156	&	5557296	&	11362.1524	&	20181.72	&	18092.8895	&	1	\\
Jiangsu	&	DM-10	&	30854.2394	&	8309369	&	28849.8414	&	54058.22	&	57690.4826	&	1	\\
Zhejiang	&	DM-11	&	17649.3612	&	10701227	&	18076.1831	&	34665.33	&	32888.3252	&	1	\\
Anhui	&	DM-12	&	15425.8327	&	4368291	&	11357.9481	&	17212.0506	&	29265.9758	&	2	\\
Fujian	&	DM-13	&	12439.9351	&	6378556	&	11185.4411	&	19701.78	&	19493.2447	&	2	\\
Jiangxi	&	DM-14	&	10774.1579	&	3857934	&	7232.9199	&	12948.88	&	14317.0176	&	2	\\
Shandong	&	DM-15	&	31255.9773	&	11101706	&	38899.2491	&	50013.2449	&	86418.6422	&	2	\\
Henan	&	DM-16	&	21449.9985	&	8811826	&	23647.1137	&	29599.31	&	50183.8533	&	1	\\
Hubei	&	DM-17	&	15578.2914	&	5980034	&	17674.6591	&	22250.45	&	33167.6494	&	2	\\
Hunan	&	DM-18	&	14523.2404	&	5674859	&	16744.0822	&	22154.2275	&	24898.3247	&	2	\\
Guangdong	&	DM-19	&	18751.4679	&	13039847	&	29144.0130	&	57067.9177	&	44558.6746	&	2	\\
Guangxi	&	DM-20	&	9808.6135	&	3579829	&	9154.5046	&	13035.1015	&	15609.2726	&	2	\\
Hainan	&	DM-21	&	2145.3794	&	900790	&	1687.9811	&	2855.54	&	3735.6048	&	3	\\
Chongqing	&	DM-22	&	8736.1666	&	3531876	&	9278.4060	&	11409.6	&	15422.0644	&	2	\\
Sichuan	&	DM-23	&	17039.9763	&	6408886	&	20574.9974	&	23872.8	&	29363.4527	&	2	\\
Guizhou	&	DM-24	&	5717.8049	&	2694970	&	9878.3789	&	6852.2	&	25350.7651	&	2	\\
Yunnan	&	DM-25	&	7831.1300	&	3926735	&	10433.6832	&	10309.47	&	20673.835	&	2	\\
Shaanxi	&	DM-26	&	12044.5492	&	4112237	&	10625.7099	&	14453.68	&	32141.9093	&	1	\\
Gansu	&	DM-27	&	5145.0277	&	2113309	&	7007.0411	&	5650.204	&	13123.5733	&	1	\\
Ningxia	&	DM-28	&	2096.8642	&	674222	&	4562.3895	&	2341.29	&	15234.0677	&	2	\\
Xinjiang	&	DM-29	&	6158.7750	&	2887710	&	11831.3864	&	7505.31	&	25516.5171	&	1\\ \hline

\end{tabularx}
 \end{table}

\medskip
\providecommand{\url}[1]{\texttt{#1}}
\providecommand{\urlprefix}{URL }


%\begin{document}
@article{Antunes,
author = {Antunes, Jorge and Hadi-Vencheh, Abdollah and Jamshidi, Ali and Tan, Yong and Wanke, Peter}, 
title = {TEA-IS: A hybrid DEA-TOPSIS approach for assessing performance and synergy in Chinese health care}, 
year = {2023}, 
publisher = {Elsevier Science Publishers B. V.}, 
address = {NLD}, 
volume = {171}, 
number = {C}, 
issn = {0167-9236}, 
url = {https://doi.org/10.1016/j.dss.2022.113916},
@book{Dantzig,
address = {Berlin, Heidelberg},
author = {Dantzig, George B and Thapa, Mukund N},
isbn = {0387948333},
publisher = {Springer-Verlag},
title = {{Linear programming 1: introduction}},
year = {1997}
}
@article{Taherdoost2023,
author = {Taherdoost, Hamed and Madanchian, Mitra},
doi = {10.3390/encyclopedia3010006},
file = {:Users/Franklin{\_}Liu/Downloads/encyclopedia-03-00006.pdf:pdf},
journal = {Encyclopedia},
number = {1},
pages = {77--87},
title = {{Multi-Criteria Decision Making (MCDM) Methods and Concepts}},
volume = {3},
year = {2023}
}
@article{Amor2023,
author = {Amor, Sarah Ben and Belaid, Fateh and Benkraiem, Ramzi and Ramdani, Boumediene and Guesmi, Khaled},
doi = {10.1007/s10479-022-04986-9},
file = {:Users/Franklin{\_}Liu/Desktop/IJITDS paper/IJITDM paper 2025-1-17/Amor et al., 2023, Multi-criteria classification, sorting, and clustering a bibliometric review and research agenda.pdf:pdf},
issn = {15729338},
journal = {Ann. Oper. Res.},
number = {2},
pages = {771--793},
publisher = {Springer US},
title = {{Multi-criteria classification, sorting, and clustering: a bibliometric review and research agenda}},
url = {https://doi.org/10.1007/s10479-022-04986-9},
volume = {325},
year = {2023}
}
@article{Dyson2001,
author = {Dyson, R G and Allen, R and Camanho, A S and Podinovski, V V and Sarrico, C S and Shale, E A},
doi = {https://doi.org/10.1016/S0377-2217(00)00149-1},
issn = {0377-2217},
journal = {Eur. J. Oper. Res.},
number = {2},
pages = {245--259},
title = {{Pitfalls and protocols in DEA}},
url = {https://www.sciencedirect.com/science/article/pii/S0377221700001491},
volume = {132},
year = {2001}
}
@article{Keenan,
author = {Keenan, Peter},
doi = {10.1080/12460125.2024.2354642},
journal = {J. Decis. Syst.},
number = {supl},
pages = {78--88},
publisher = {Taylor $\backslash${\&} Francis},
title = {{A scientometric analysis of multicriteria decision-making research}},
url = {https://doi.org/10.1080/12460125.2024.2354642},
volume = {33},
year = {2024}
}
@article{Danielson2014,
author = {Danielson, M and Ekenberg, L and He, Y},
journal = {Decis. Anal.},
url = {https://doi.org/10.1287/deca.2013.0289},
number = {1},
pages = {21--26},
title = {{Augmenting ordinal methods of attribute weight approximation.}},
volume = {11},
year = {2014}
}
@article{Amin:2013,
author = {Amin, Gholam R. and Emrouznejad, Ali},
doi = {10.1007/s00170-012-4280-3},
file = {:Users/Franklin{\_}Liu/Downloads/2013-J-IJAMT-65-Journal-PrePub-Published-2.pdf:pdf},
journal = {Int. J. Adv. Manuf. Technol.},
number = {9-12},
pages = {1567--1572},
title = {{A new DEA model for technology selection in the presence of ordinal data}},
volume = {65},
year = {2013}
}
@article{Zarrin,
author = {Zarrin, Mansour and Schoenfelder, Jan and Brunner, Jens O},
doi = {10.1007/s10729-022-09590-8},
issn = {1572-9389},
journal = {Health Care Manag. Sci.},
number = {3},
pages = {406--425},
title = {{Homogeneity and Best Practice Analyses in Hospital Performance Management: An Analytical Framework}},
url = {https://doi.org/10.1007/s10729-022-09590-8},
volume = {25},
year = {2022}
}
@misc{Liu2023,
author = {Liu, Fuh-Hwa Franklin and Shih, Su-Chuan},
title = {{Data envelopment analysis models or the virtual gap analysis model: Which should be used for identifying the best benchmark for each unit in a group?}},
eprint={2306.11224},
      archivePrefix={arXiv},
      primaryClass={math.OC},
      url={https://arxiv.org/abs/2306.11224},
year = {2023}
}
@article{Amin:2022,
author = {Amin, Gholam R and El-Temtamy, Osama and Garas, Samy},
journal = {Abacus},
title = {{Audit Risk Evaluation Using Data Envelopment Analysis with Ordinal Data}},
url = {https://api.semanticscholar.org/CorpusID:248146047},
year = {2022}
}
@article{Sahoo,
author = {Sahoo, Sushil Kumar and Goswami, Shankha Shubhra},
doi = {10.31181/dma1120237},
file = {:Users/Franklin{\_}Liu/Downloads/DecisionMakingAdvancesDMA{\_}ScientificOasis1.pdf:pdf},
journal = {Decis. Mak. Adv.},
number = {1},
pages = {25--48},
title = {{A Comprehensive Review of Multiple Criteria Decision-Making (MCDM) Methods: Advancements, Applications, and Future Directions}},
volume = {1},
year = {2023}
}
@article{Ebrahimi2020,
author = {Ebrahimi, B. and Dellnitz, A. and Kleine, A. and Tavana, M},
doi = {10.1016/j.eswa.2020.113835},
file = {:Users/Franklin{\_}Liu/Downloads/Ebrahimi et al., 2021, A novel method for solving data envelopment analysis problems with weak ordinal data using robust measures.pdf:pdf},
issn = {09574174},
journal = {Expert Syst. Appl.},
number = {August},
title = {{A novel method for solving data envelopment analysis problems with weak ordinal data using robust measures}},
volume = {164},
year = {2021}
}
@article{Power2019,
author = {Power, D and Heavin, C and Keenan, P},
doi = {10.1080/12460125.2019.1631683},
journal = {J. Decis. Syst.},
pages = {1--18},
title = {{Decision systems redux}},
volume = {28},
year = {2019}
}
@article{Ebrahimi:2024,
author = {Ebrahimi, B. and Tesic, D.},
doi = {10.31181/sa22202425},
file = {:Users/Franklin{\_}Liu/Downloads/Ebrahimi, 2024, A DEA based Performance Measurement Approach with Weak Ordinal Data Paper Type Original Article A DEA based Performance.pdf:pdf},
Journal = {Syst. Anal.}, 
number = {2},
pages = {229-242},
title = {{A DEA based Performance Measurement Approach with Weak Ordinal Data Paper Type : Original Article A DEA based Performance Measurement Approach with Weak Ordinal Data}},
volume = {2},
year = {2024}
}
@article{Mergoni,
author = {Mergoni, Anna and {De Witte}, Kristof},
doi = {https://doi.org/10.1111/itor.13012},
journal = {Int. Trans. Oper. Res.},
number = {3},
pages = {1337--1359},
title = {{Policy evaluation and efficiency: a systematic literature review}},
url = {https://onlinelibrary.wiley.com/doi/abs/10.1111/itor.13012},
volume = {29},
year = {2022}
}
@inproceedings{Liu2015,
address = {Dubai},
author = {Liu, Fuh-Hwa Franklin and Hwang, Yang-Cheng},
booktitle = {Int. Res. Conf. Sci. Manag. Eng. (IRCSME 2015, Dubai)},
pages = {SME14025.},
title = {{Virtual-gap measurement (VGM) for assessing a set of units}},
year = {2015}
}
@article{Tone,
author = {Tone, Kaoru},
doi = {https://doi.org/10.1016/S0377-2217(99)00407-5},
issn = {0377-2217},
journal = {Eur. J. Oper. Res.},
number = {3},
pages = {498--509},
title = {{A slacks-based measure of efficiency in data envelopment analysis}},
volume = {130},
year = {2001}
}
@article{Halicka,
author = {Halick{\'{a}}, Margar{\'{e}}ta and Trnovsk{\'{a}}, M{\'{a}}ria},
doi = {10.1016/j.ejor.2020.07.019},
issn = {03772217},
journal = {Eur. J. Oper. Res.},
number = {2},
pages = {611--627},
title = {{A unified approach to non-radial graph models in data envelopment analysis: common features, geometry, and duality}},
volume = {289},
year = {2021}
}
@article{Emrouznejad,
author = {Emrouznejad, A. and Brzezicki, L. and Lu, C.},
title = {{Development and Evolution of Slacks-Based Measure Models in Data Envelopment Analysis: A Comprehensive Review of the Literature.}},
journal ={J Econ Surv. },
doi ={https://doi.org/10.1111/joes.12682},
year = {2025}
}
@incollection{Danielson2022,
address = {Cham},
author = {Danielson, Mats and Ekenberg, Love},
booktitle = {Multicriteria Optim. Model. Risk, Reliab. Maint. Decis. Anal. Recent Adv.},
editor = {Almeida, A T and  Ekenberg, L and Scarf, P and Zio, E and Zuo, M J},
doi = {https://10.1007/978-3-030-89647-8-2},
pages = {29--40},
publisher = {Springer International Publishing},
title = {Comparing Cardinal and Ordinal Ranking in MCDM Methods},
url = {https://doi.org/10.1007/978-3-030-89647-8{\_}2},
year = {2022}
}
@article{Aleskerov,
author = {Aleskerov, Fuad and Petrushchenko, Vsevolod},
doi = {10.1142/S021962201550042X},
issn = {02196220},
journal = {Int. J. Inf. Technol. Decis. Mak.},
number = {1},
pages = {5--22},
title = {{DEA by sequential exclusion of alternatives in heterogeneous samples}},
volume = {15},
year = {2016}
}
@article{Liu2017b,
author = {Liu, Fuh-Hwa Franklin and Liu, Yu-Cheng},
doi = {10.1016/j.cie.2017.06.010},
issn = {03608352},
journal = {Comput. Ind. Eng.},
pages = {550--559},
title = {{A methodology to assess the supply chain performance based on gap-based measures}},
volume = {110},
year = {2017}
}
@article{Heavin,
author = {Heavin, Ciara and Power, Daniel J},
doi = {10.1080/12460125.2018.1468697},
journal = {J. Decis. Syst.},
number = {sup1},
pages = {38--45},
publisher = {Taylor $\backslash${\&} Francis},
title = {{Challenges for digital transformation – towards a conceptual decision support guide for managers}},
url = {https://doi.org/10.1080/12460125.2018.1468697},
volume = {27},
year = {2018}
}
@article{Chen,
author = {Chen, Y and Cook, WD and Du, J and Hu, H-H and Zhu, J},
doi = {10.1007/s10479-015-1827-3},
file = {:Users/Franklin{\_}Liu/Downloads/Chen et al., 2017, Bounded and discrete data and Likert scales in data envelopment analysis application to regional energy efficiency in.pdf:pdf},
issn = {15729338},
journal = {Ann. Oper. Res.},
number = {1-2},
pages = {347--366},
publisher = {Springer US},
title = {{Bounded and discrete data and Likert scales in data envelopment analysis: application to regional energy efficiency in China}},
url = {http://dx.doi.org/10.1007/s10479-015-1827-3},
volume = {255},
year = {2017}
}
@article{2016LiW,
author = {Li, W.-H. and Liang, L. and Cook, W. and Zhu, J.},
doi = {10.1016/j.ejor.2016.04.063},
issn = {03772217},
journal = {Eur. J. Oper. Res.},
number = {3},
pages = {946--956},
publisher = {Elsevier B.V.},
title = {{DEA models for non-homogeneous DMUs with different input configurations}},
url = {http://dx.doi.org/10.1016/j.ejor.2016.04.063},
volume = {254},
year = {2016}
}
@article{Liu2017a,
author = {Liu, Fuh-Hwa Franklin and Liu, Yu-Cheng},
doi = {10.1155/2017/3060342},
issn = {15635147},
journal = {Math. Probl. Eng.},
title = {{Procedure to Solve Network DEA Based on a Virtual Gap Measurement Model}},
volume = {2017},
year = {2017}
}
@article{Zhu,
 author = {Zhu, WW and Yu, Y and Sun, PP},
journal = {Eur. J. Oper. Res.},
number = {1},
pages = {99--110},
title = {{Data envelopment analysis cross-like efficiency model for non-homogeneous decision-making units: The case of United States companies' low-carbon investment to attain corporate sustainability}},
url = {https://www.sciencedirect.com/science/article/pii/S0377221717307191},
volume = {269},
year = {2018}
}
@thesis{LinZY2024,
author = {Lin, Z-Y},
title = {{Analysis of product mix strategy for camping supplies stores}},
school = {Providence University, Taiwan},
Type = {{Unpublished Master Thesis (in Chinese), advisor: Professor Shih S-C}},
year = {2024}
}
@article{Mardani2015,
author = {Mardani, Abbas and Jusoh, Ahmad and Nor, Khalil M.D. and Khalifah, Zainab and Zakwan, Norhayati and Valipour, Alireza},
doi = {10.1080/1331677X.2015.1075139},
file = {:Users/Franklin{\_}Liu/Downloads/mardani2015.pdf:pdf},
issn = {1331677X},
journal = {Econ. Res. Istraz. },
number = {1},
pages = {516--571},
publisher = {Routledge},
title = {{Multiple criteria decision-making techniques and their applications - A review of the literature from 2000 to 2014}},
url = {http://dx.doi.org/10.1080/1331677X.2015.1075139},
volume = {28},
year = {2015}
}
@misc{Liu2025,
author = {Liu, Fuh-Hwa Franklin and Shih, Su-Chuan},
pages = {28},
title = {{Algorithms for Multicriteria Decision-Making and Efficiency Analysis Problems.}},
eprint={2406.06090},
      archivePrefix={arXiv},
      primaryClass={math.FA},
      url={https://arxiv.org/abs/2406.06090},
year ={2025}
}
@article{Ceballos,
author = {Ceballos, B and Lamata, M. T.  and Pelta, D A},
doi = {10.1007/s13748-016-0093-1},
file = {:Users/Franklin{\_}Liu/Downloads/s13748-016-0093-1.pdf:pdf},
issn = {21926360},
journal = {Prog. Artif. Intell.},
number = {4},
pages = {315--322},
publisher = {Springer Berlin Heidelberg},
title = {{A comparative analysis of multi-criteria decision-making methods}},
volume = {5},
year = {2016}
}
@article{Emrouznejad,
author = {Emrouznejad, Ali and Brzezicki, L and Lu, C},
doi = {https://doi.org/10.1111/joes.12682},
journal = {J. Econ. Surv.},
number = {n/a},
title = {{Development and Evolution of Slacks-Based Measure Models in Data Envelopment Analysis: A Comprehensive Review of the Literature}},
year ={2025}
url = {https://onlinelibrary.wiley.com/doi/abs/10.1111/joes.12682},
volume = {n/a}
}
@article{Brown,
author = {Brown, Rayna},
doi = {https://doi.org/10.1016/j.ejor.2005.03.025},
issn = {0377-2217},
journal = {Eur. J. Oper. Res.},
number = {2},
pages = {1100--1116},
title = {{Mismanagement or mismeasurement? Pitfalls and protocols for DEA studies in the financial services sector}},
url = {https://www.sciencedirect.com/science/article/pii/S0377221705003498},
volume = {174},
year = {2006}
}
@book{Power2017,
author = {Power, D and Heavin, C},
isbn = {9781637423325},
publisher = {Business Expert Press},
title = {{Decision Support, Analytics, and Business Intelligence, Third Edition}},
url = {https://books.google.com.tw/books?id=dpXozgEACAAJ},
year = {2017}
}
@thesis{JCLi2022,
  author  = {Li, J-C},
  title = {Evaluating the internet word-of-mouth performance of the hotel industry through online travel reviews},
Type = {Unpublished Master Thesis (in Chinese), advisor: Professor Shih S-C},
school = {Providence University, Taiwan},
year = {2022},
}
@book{Cooper,
   author = {Cooper, William and Seiford, Lawrence M. and Tone, Kaoru},
  title = {{DATA ENVELOPMENT ANALYSIS A Comprehensive Text with Models, Applications, References and DEA-Solver Software Second Edition}},
 publisher = {Springer Science+Business Media, LLC},
address = {New York},
 year = {2007} 
}
@article{Toloo,
author = {Toloo, Mehdi and Nalchigar, Soroosh},
doi = {10.1016/j.eswa.2011.05.008},
file = {:Users/Franklin{\_}Liu/Downloads/1-s2.0-S0957417411007913-main.pdf:pdf},
issn = {09574174},
journal = {Expert Syst. Appl.},
number = {12},
pages = {14726--14731},
publisher = {Elsevier Ltd},
title = {{A new DEA method for supplier selection in presence of both cardinal and ordinal data}},
url = {http://dx.doi.org/10.1016/j.eswa.2011.05.008},
volume = {38},
year = {2011}
}
%\end{document}


\begin{thebibliography}{38}
\bibitem[{Aleskerov and Petrushchenko(2016)}] {Aleskerov} F. Aleskerov and V. Petrushchenko, DEA by sequential exclusion of alternatives in heterogeneous samples, \textit{IJITDM.} \textbf{15}(1)(2016) 5-22.

\bibitem[{Amin and Emrouznejad(2013)}] {Amin:2013} G.R. Amin and A. Emrouznejad, A new DEA model for technology selection in the presence of ordinal data, \textit{Int. J. Adv. Manuf. Technol.} \textbf{65} (2013) 1567-1572.

\bibitem[{Amin et~al.(2022)}] {Amin:2022} G.R. Amin, O. El-Temtamy, and S. Garas, Audit Risk Evaluation Using Data Envelopment Analysis with Ordinal Data, \textit{Abacus} \textbf{58}(3)(2022) 589–602.

\bibitem[{Amor et~al.(2022)}]{Amor} S.B. Amor, F. Belaid, R. Benkraiem, B. Ramdani, and K. Guesmi, Multicriteria classification, sorting, and clustering: A bibliometric review and research agenda, \textit{A. Oper. Res.} \textbf{325}(2)(2022) 1-23.

\bibitem[{Antunes et~al.(2023)}] {Antunes} J. Antunes, A. Hadi-Vencheh, A. Jamshidi, Y. Tan, and P. Wanke, TEA-IS: A hybrid DEA-TOPSIS approach for assessing performance and synergy in Chinese health care, \textit{Decis. Support Syst.}, \textbf{171} (2023), 113916.

\bibitem[{Ban(2011)}]{Ban} O.I. Ban, Fuzzy multicriteria decision making method applied to selection of the best touristic destinations, \textit{IJMMMAS} \textbf{2} (5) 2011 264-271.

\bibitem[{Bazaraa et~al.(2011)}] {Bazaraa} M.S. Bazaraa, J.J. Jarvis, and H.D. Sherali, \textit{Linear Programming and Network Flows}, 4th Edition (John Wiley \& Sons, 2011).

\bibitem[{Černevičienė and  Kabašinskas(2022)}]{Cernev} J. Černevičienė and A. Kabašinskas, Review of multicriteria decision-making methods in finance using explainable artificial intelligence, \textit{Front. Artif. Intell.} \textbf{5} (2022).

\bibitem[{Chen et al.(2017)}]{Chen} Y. Chen, W.D. Cook, J. Du, H. Hu, and J. Zhu, Bounded and discrete data and Likert scales in data envelopment analysis: Application to regional energy efficiency in China, \textit{A. Oper. Res.} \textbf{255} (2017) 47–366. 

\bibitem[{Danielson and Ekenberg(2022)}]{Danielson} M. Danielson and L. Ekenberg, Comparing cardinal and ordinal ranking in MCDM Methods. In:  A.T. Almeida, L. Ekenberg, P. Scarf, E. Zio, M.J. Zuo (eds) \textit{Multicriteria and Optimization Models for Risk, Reliability, and Maintenance Decision Analysis}, International Series in Operations Research and Management Science, \textbf{321} (Springer Cham. 2022).  

\bibitem[{Danielson et~al.(2014)}]{Danielson2014} M. Danielson, L. Ekenberg, and Y. He, Augmenting ordinal methods of attribute weight approximation, \textit{Decis. Anal.} \textbf{11}(1) (2014) 21-26.

\bibitem[{Dyson et~al.(2001)}]{Dyson} R.G. Dyson, R. Allen, A.S. Camanho, V.V. Podinovski, C.S. Sarrico, and E.A. Shale, (2001). Pitfalls and protocols in DEA. \textit{Eur. J. Oper. Res.} \textbf{132}(2), 245–259. 

\bibitem[{Ebrahimi and Tesic(2024)}]{Ebrahimi:2024} B. Ebrahimi and D. Tesic, A DEA-based performance measurement approach with weak ordinal data, \textit{Syst. Anal.} \textbf{2}(2)(2024) 229-242.

\bibitem[{Ebrahimi et~al.(2021)}]{Ebrahimi:2020} B. Ebrahimi, A. Dellnitz, A. Kleine, and M. Tavana, 2021. A Novel method for solving data envelopment analysis problems with weak ordinal data using robust measures, \textit{Expert Syst. Appl.} \textbf{164}, 113835. 

\bibitem[{Emrouznejad et~al.(2025)}] {Emrouznejad} A. Emrouznejad, L. Brzezicki, and C. Lu, Development and evolution of slacks-based measure models in data envelopment analysis: A comprehensive review of the literature, \textit{J. Econ. Surv.} (2025).  

\bibitem[{Halická and Trnovská(2021)}]{Halicka} M. Halická and M. Trnovská, A unified approach to non-radial graph models in data envelopment analysis: Common features, geometry, and duality, \textit{Eur. J. Oper. Res.} \textbf{289}(2) (2021) 611–627.

\bibitem [{Hansen and F Ombler(2008)}] {Hansen} P. Hansen and F. Ombler, A new method for scoring multi-attribute value models using pairwise rankings of alternatives, \textit{JMCDA} \textbf{15} (2008) 87-107.

\bibitem[{Heavin and Power(2018)}]{Heavin} C. Heavin and D.J. Power, Challenges for digital transformation – toward a conceptual decision support guide for managers, \textit{JDS.} \textbf{27}(sup1) (2018) 38–45.

 \bibitem[{Kumbhakar et~al.(2022)}]{Kumbhakar} S.C. Kumbhakar, C.F. Parmeter, and V. Zelenyuk, Stochastic Frontier Analysis: Foundations and Advances I. In: Ray, S.C., Chambers, R.G., and Kumbhakar, S.C. (eds) Handbook of Production Economics. (Springer, Singapore 2022)

\bibitem[{Keenan(2024)}]{Keenan} P. Keenan, A scientometric analysis of multicriteria decision-making research, \textit{JDS.} \textbf{33}(sup1)(2024) 78-88. 

\bibitem[{Li et~al.(2016)}] {2016LiW} W.H. Li, L. Liang, W.D. Cook, and J. Zhu, DEA models for non-homogeneous DMUs with different input configurations, \textit{Eur. Oper. Res.} \textbf{254}(3) (2016) 946-956.

\bibitem[{Li(2022)}] {JCLi2022} J.C. Li, Evaluating the internet word-of-mouth performance of the hotel industry
through online travel reviews, Master thesis (in Chinese), Providence University, Taiwan 2022.

\bibitem[{Lin(2024)}] {LinZY2024} Z.Y. Lin, Analysis of product mix strategy for camping supplies stores. Master thesis (in Chinese), Providence University, Taiwan 2024.

\bibitem[{Liu and Hwang(2015)}] {Liu2015} F.F.-H. Liu and Y.-C Hwang, Virtual-gap measurement (VGM) for assessing a set of units. Int. Res. Conf. Sci. Manag. Eng. (IRCSME 2015, Dubai), SME14025.

\bibitem[{Liu and Liu(2017[\natexlab{a}])}] {Liu2017a}  F.F.-H. Liu and Y.-C. Liu,  Procedure to Solve Network DEA Based on a Virtual Gap Measurement Model,  \textit{Math. Probl. Eng.} (2017a) 1-13.

\bibitem[{Liu and Liu(2017[\natexlab{b}])}] {Liu2017b}  F.F.-H. Liu and Y.-C. Liu,  A methodology to assess the supply chain performance based on gap-based measures, \textit{Comput. Ind. Eng.} \textbf{110}(2017b) 550-559.

\bibitem[{Liu and Shih(2023)}]{Liu2023}  F.F.-H. Liu and S.-C Shih, Data envelopment analysis models or the virtual gap analysis model: Which should be used for identifying the best benchmark for each unit in a group? (2023) \textit{Digital preprint, arxiv.org/abs/2306.11224.}

\bibitem[{Liu and Shih(2025)}]{Liu2025}  F.F.-H. Liu and S.-C. Shih, Algorithms for Multi-Criteria Decision-Making and Efficiency Analysis Problems, \textit{IJITDM} (Accepted 2025) doi.org/10.1142/S0219622025500646.

\bibitem[{Mardani et~al.(2015)}]{Mardani} A. Mardani, A. Jusoh, H. M.D. Nor, Z. Khalifah, N. Zakwan, and A. Valipour, Multiple criteria decision-making techniques and their applications – a review of the literature from 2000 to 2014, \textit{Econ. Res.-Ekon. Istraz.} \textbf{28}(1)(2015) 516–571. 

\bibitem[{Mergoni and Witte(2021)}] {Mergoni} A. Mergoni and K.D. Witte, Policy evaluation and efficiency: A systematic literature review, \textit{Int. Trans. Oper. Res.} \textbf{29}(3)(2021) 1337-1359.

\bibitem[{Power et~al.(2019)}]{Power2019} D.J. Power, C. Heavin, and P. Keenan, Decision systems redux, \textit{JDS.} \textbf{28}(1)(2019) 1–18. 

\bibitem[{Sahoo and Goswami(2023)}] {Sahoo} S. Sahoo and S. Goswami, A comprehensive review of multiple conditions decision-making (MCDM) Methods: Advancements, applications, and future directions, \textit{DMA.} \textbf{1}(1)(2023) 25–48.

\bibitem[{Sickles and Zelenyuk(2019)}]{Sickle} R.C. Sickles and V. Zelenyuk, \textit{Measurement of Productivity and Efficiency: Theory and Practice,} (Cambridge University Press, 2019).

\bibitem [{Scholz et~al(2016)}] {Scholz}
V. Scholz, A. Kirbyshire, and N. Simister, Shedding light on causal recipes for development research uptake: Applying Qualitative Comparative Analysis to understand reasons for research uptake, INTRAC and CDKN, April 2016.

\bibitem[{Taherdoost and Madanchian(2023)}]{Taherdoost} H. Taherdoost and M. Madanchian, Multicriteria decision making (MCDM) methods and concepts, \textit{Encyclopedia} \textbf{3}(1)(2023) 77-87.

\bibitem[Theodoridis et~al.(2008)] {Theodoridis} A.M. Theodoridis and A. Psychoudakis, Efficiency measurement in Greek dairy farms: Stochastic frontier vs. data envelopment analysis, \textit{IJESAR}, \textbf{1}(2008)(2) 53-66.

\bibitem[{Toloo and Nalchigar(2011)}] {Toloo} M. Toloo and S. Nalchigar, A new DEA method for supplier selection in the presence of both cardinal and ordinal data, \textit{Expert Syst. Appl.} \textbf{38}(12)(2011) 147–14731.

\bibitem[{Volaric et~al.(2014)}]{Volaric} T. Volaric, E. Brajković, and T. Sjekavica, Integration of FAHP and TOPSIS methods for the selection of appropriate multimedia application for learning and teaching, \textit{IJMMMAS} \textbf{8}(2014) 224-232.

\bibitem[{Zarrin et~al.(2022)}] {Zarrin} M. Zarrin, J. Schoenfelder, and J.O. Brunner, Homogeneity and best practice analyses in hospital performance management: An analytical framework, \textit{Health Care Manag Sci.}  \textbf{25}(3)(2022) 406–425. 

\bibitem[{Zhu et~al.(2018)}]{Zhu} W.W. Zhu, Y. Yu, and P.P. Sun, Data envelopment analysis cross-like efficiency model for non-homogeneous decision-making units: The case of United States companies' low-carbon investment to attain corporate sustainability, \textit{Eur. J. Oper. Res.} \textbf{269}(1)(2018) 99–110.
\end{thebibliography}
\end{document}